\documentclass[10pt,journal,cspaper,compsoc]{IEEEtran}
\usepackage[breaklinks=true,colorlinks,citecolor=blue,linkcolor=blue,urlcolor=blue]{hyperref}
\usepackage{url}            
\usepackage{booktabs}       
\usepackage{amsfonts}       
\usepackage{nicefrac}       
\usepackage{microtype}      
\usepackage{xcolor}         
\usepackage{subfigure}
\usepackage{times}
\usepackage{graphicx}
\usepackage{amsmath}
\usepackage{amssymb}
\usepackage[group-separator={,},group-minimum-digits={3}]{siunitx}
\usepackage{color}
\usepackage{xcolor}
\usepackage{multirow}
\usepackage{overpic}

\usepackage[nocompress]{cite} 
\usepackage{ragged2e}

\usepackage{cancel}  
\usepackage[normalem]{ulem}

\newcommand{\figref}[1]{Figure~\ref{#1}}%
\newcommand{\tabref}[1]{Table~\ref{#1}}%
\newcommand{\secref}[1]{Section~\ref{#1}}
\newcommand{\myPara}[1]{\vspace{.15in} \noindent\textbf{#1}}
\newcommand{\ourdata}{{ImageNet-S}}

\def\ie{\emph{i.e.,~}}
\def\eg{\emph{e.g.,~}}
\def\etc{\emph{etc}}
\def\etal{{\em et al.}}
\def\Real{\mathbb{R}}
\def\Pooling{\mathbb{P}}

\def\OursP{\textbf{PASS$_p$}}
\def\OursS{\textbf{PASS$_s$}}

\def\SimCLR{SimCLR\cite{chen2020simple}}
\def\BYOL{BYOL\cite{jean-bastien2020bootstrap}}
\def\MoCovT{MoCov2\cite{chen2020mocov2, He_2020_CVPR}}
\def\DenseCL{DenseCL\cite{wang2020DenseCL}}
\def\AdCo{AdCo\cite{hu2020adco}}
\def\PCL{PCL\cite{PCL}}
\def\SwAV{SwAV\cite{caron2020unsupervised}}
\def\PixelPro{PixelPro\cite{xie2020propagate}}
\def\SimSiam{SimSiam\cite{chen2020exploring}}

\usepackage{silence}
\graphicspath{{./figures/}{./figures/datasets/}}

\newcommand{\addImg}[2]{\includegraphics[width=#1\linewidth]{#2}}

\begin{document}

\title{Large-scale Unsupervised Semantic Segmentation}

\author{%
	Shanghua Gao, Zhong-Yu Li, Ming-Hsuan Yang, Ming-Ming Cheng, Junwei Han, Philip Torr
	\IEEEcompsocitemizethanks{
		\IEEEcompsocthanksitem S. Gao, Z.Y. Li, and M.M. Cheng are with CS,
		Nankai University. M.M. Cheng is the corresponding author (cmm@nankai.edu.cn).
		\IEEEcompsocthanksitem M.H. Yang is with UC Merced.
		\IEEEcompsocthanksitem J. Han is with Northwestern Polytechnical University.
		\IEEEcompsocthanksitem P. Torr is with Oxford University.
	}
}

\markboth{IEEE Transactions on Pattern Analysis and Machine Intelligence}
{Gao \MakeLowercase{\textit{et al.}}:
	Large-scale Unsupervised Semantic Segmentation}

\IEEEtitleabstractindextext{
	\begin{abstract}
		\justifying
		Empowered by large datasets, e.g., ImageNet, unsupervised learning on large-scale data has enabled significant advances
		for classification tasks.
		However, whether the large-scale unsupervised semantic segmentation
		can be achieved remains unknown.
		There are two major challenges:
		i) we need a large-scale benchmark for assessing algorithms;
		ii) we need to develop methods to simultaneously learn category and shape representation in an unsupervised manner.
		In this work, we propose a new problem of \textbf{l}arge-scale
		\textbf{u}nsupervised \textbf{s}emantic \textbf{s}egmentation (LUSS)
		with a newly created benchmark dataset to help the research progress.
		Building on the ImageNet dataset,
		we propose the \ourdata~dataset with 1.2 million training images
		and 50k high-quality semantic segmentation annotations for evaluation.
		Our benchmark has a high data diversity and a clear task objective.
		We also present a simple yet effective method that works
		surprisingly well for LUSS.
		In addition, we benchmark related un/weakly/fully supervised methods accordingly,
		identifying the challenges and possible directions of LUSS.
		The benchmark and source code is publicly available at
		\url{https://github.com/LUSSeg}.
	\end{abstract}
	\begin{IEEEkeywords}
		large-scale, unsupervised, semantic segmentation, self-supervised, ImageNet
	\end{IEEEkeywords}
}

\maketitle
\IEEEdisplaynontitleabstractindextext
\IEEEpeerreviewmaketitle

\IEEEraisesectionheading{\section{Introduction}\label{sec:introduction}}
\IEEEPARstart{S}{emantic} segmentation
\cite{long2015fully,yu2015multi,ronneberger2015u,
	badrinarayanan2017segnet,chen2017deeplab,Zhen_2020_CVPR},
aiming to label image pixels with category information,
has drawn much research attention.
Due to the inherent challenges of this task,
most efforts focus on semantic segmentation under environments with
limited diversity~\cite{Cordts2016Cityscapes,zhou2018semantic,yu2020bdd100k}
and data scale~\cite{Everingham15,caesar2018cvpr}.
For instance, the PASCAL VOC segmentation dataset only contains about 2k images,
while the BDD100K~\cite{yu2020bdd100k} focuses on road scenes.
Numerous approaches have achieved impressive results in these
restricted environments~\cite{Chen_2018_ECCV,fan2021rethinking,
	wang2021crossdataset,zhu2021learning,zhao2019region,
	liu2020efficientfcn,li2020improving,liu2020learning,yuan2021segmentation}.
Significantly scaling up the problem often results in research domain adaptation,
\eg from PASCAL VOC~\cite{Everingham15} to
ImageNet~\cite{russakovsky2015imagenet}.
This motivates us to consider a far more challenging problem:
is semantic segmentation possible for large-scale real-world
environments with a wide diversity?

\begin{figure}[t]
	\centering
	\addImg{1}{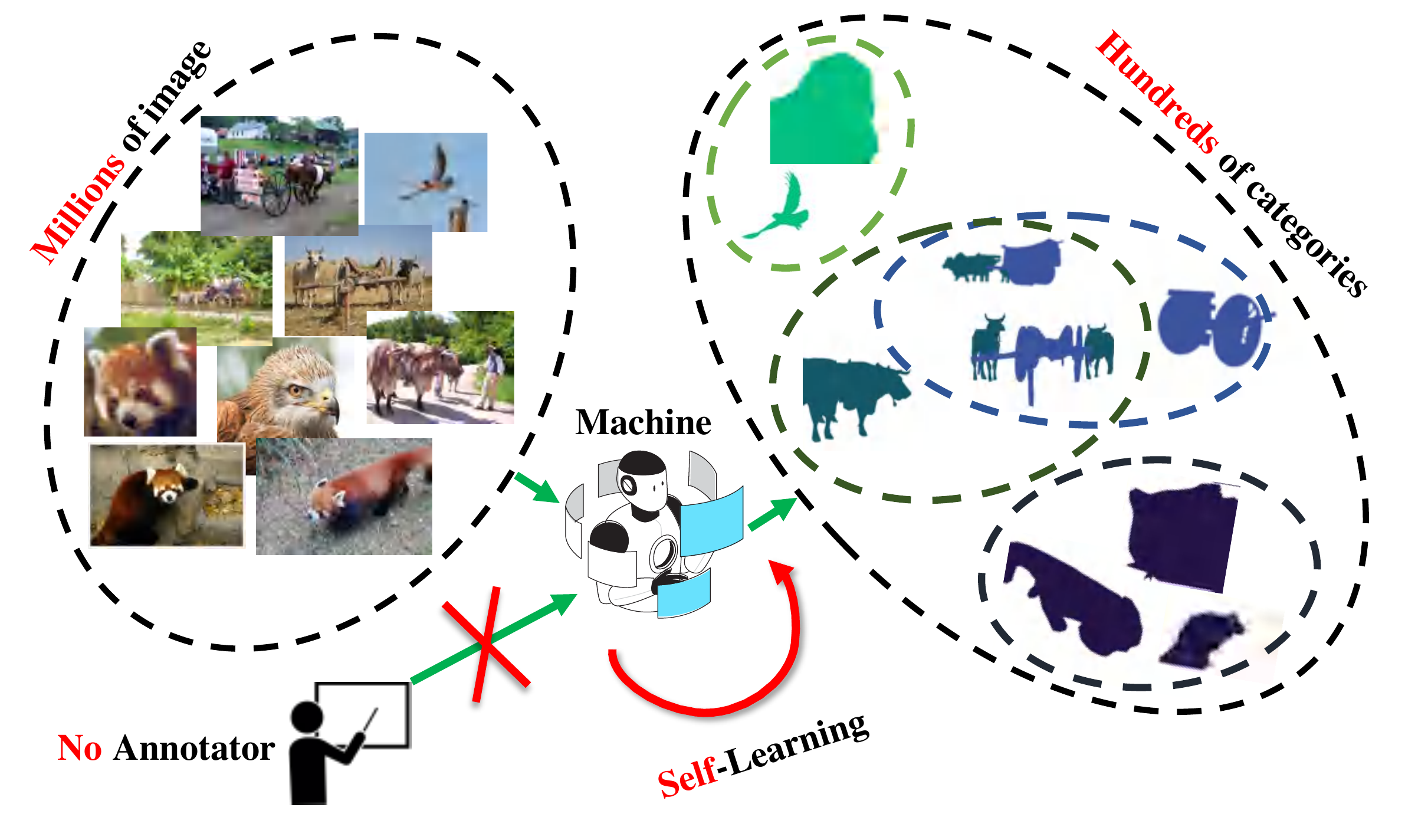} \\
	\vspace{-15pt}
	\caption{The \textbf{L}arge-scale \textbf{U}nsupervised
		\textbf{S}emantic \textbf{S}egmentation (LUSS) task aims to assign labels from \textbf{hundreds} of categories to pixels from \textbf{millions} of images without the help of human annotation.
		The model learns to conduct semantic segmentation with \textbf{Self-Learning}.  }
	\label{fig:intro}
\end{figure}

However, due to the huge data scale and privacy issues,
annotating images with pixel-level human annotations
or even image-level labels is extremely expensive.
Lacking sufficient benchmark data limits the large-scale semantic segmentation.
When trained with millions or even billions of images,
\eg ImageNet~\cite{russakovsky2015imagenet}, JFT-300M~\cite{sun2017revisiting},
and Instagram-1B~\cite{mahajan2018exploring},
unsupervised learning of classification model has recently shown
a comparable ability to supervised training
\cite{tian2020contrastive,He_2020_CVPR,chen2020simple}.
To facilitate real-world semantic segmentation,
we propose a new problem: \textbf{L}arge-scale \textbf{U}nsupervised
\textbf{S}emantic \textbf{S}egmentation (LUSS).
The LUSS task aims to assign labels to pixels from
large-scale data without human-annotation supervision,
as shown in~\figref{fig:intro}.
Many challenges, \eg simultaneously shape and category representations learning
and unsupervised semantic clustering of large amount of data,
need to be tackled to achieve this goal.
Specifically,
we need to extract semantic representations with category and shape features.
Category-related representations are required to distinguish different classes,
and shape-related representations, \eg objectness, boundary,
are the essential pixel-level cues for semantic segmentation.
The coexistence of two representations is vital to LUSS
because conflict representations might cause incorrect
semantic segmentation results.
Generating categories from large-scale data requires robust and
efficient semantic clustering algorithms.
Assigning labels to pixels requires the distinction between related
and unrelated semantic areas.
Solving these challenges for LUSS could also facilitate many related tasks.
For example, the learned shape representations from LUSS can be utilized
as the pre-training for pixel-level downstream tasks,
\eg semantic segmentation~\cite{chen2017deeplab,Chen_2018_ECCV} and
instance segmentation~\cite{he2017mask}
under restricted data scale and diversity.
Also, fine-tuning LUSS models in the semi-supervised setting facilitates
the real-world application
where a small part of large-scale data is human-labeled.

We propose a benchmark for LUSS task with high-diversity large-scale data,
as well as new sufficient evaluation protocols taking
into account different perspectives
for the LUSS task.
Large-scale data with sufficient diversity bring challenges to LUSS,
but it also provides the source for obtaining extensive representation cues.
Due to insufficient data,
a few unsupervised segmentation methods
~\cite{ouali2020autoregressive,zhan2018mix,hwang2019segsort,van2020unsuperv}
mainly
deal with small data with limited number of categories and small diversity,
thus are not suitable for the LUSS task.
We present a large-scale benchmark dataset for the LUSS task, namely \ourdata,
based on the commonly used ImageNet dataset~\cite{russakovsky2015imagenet}
in category representation learning works.
We remove the unsegmentable categories, \eg bookshop,
and utilize \num{ 919 } categories with about 1.2 million images
in ImageNet for training.
Then we annotate 40k images in the validation set of ImageNet
with precise pixel-level semantic segmentation masks for LUSS evaluation.
We also annotate about 9k images in the training set to allow more
comprehensive evaluation protocols and exploration of future applications.
Based on the more precise re-annotated image-level labels in~\cite{beyer2020we},
we enable ImageNet with multiple categories within one image.
The \ourdata~dataset provides large-scale and high-diversity data for
fairly LUSS training and sufficient evaluation.

We then present a new method for the LUSS task,
including unsupervised representation learning, label generation,
and fine-tuning steps.
For unsupervised representation learning, we propose
1) a non-contrastive pixel-to-pixel representation alignment strategy
to enhance the pixel-level shape representation without
hurting the instance-level category representation.
2) a deep-to-shallow supervision strategy to enhance the
representation quality of the network mid-level features.
The learned representation guarantees the coexistence of shape and category information.
We propose a pixel-attention scheme to highlight meaningful semantic regions for label generation,
facilitating efficient pixel-level label generation and fine-tuning
under a large data scale.
Based on the proposed method and \ourdata~dataset,
we study the relation between the proposed LUSS task and some related works
(\ie unsupervised learning~\cite{He_2020_CVPR,chen2020simple,caron2020unsupervised,jean-bastien2020bootstrap,xie2020propagate},
weakly supervised semantic segmentation
\cite{Wang_2020_CVPR,Chang_2020_CVPR,lee2021antiadversarially},
and transfer learning on downstream tasks~\cite{he2017mask,chen2017deeplab})
and identify the challenges and possible directions of LUSS.
In this work, we make two main contributions:
\begin{itemize}
	\item We propose a new large-scale unsupervised semantic segmentation problem,
	      the \ourdata~dataset with nearly 50k pixel-level annotated images,
	      919 categories, and multiple evaluation protocols.
	\item We present a novel LUSS method containing enhanced representation learning strategies and pixel-attention scheme, and we benchmark related works for LUSS.
\end{itemize}


\section{Related works}

\subsection{Unsupervised Segmentation}
Before the recent advances in deep learning,
a plethora of approaches have been developed to segment objects
with non-parametric methods (\eg
label transfer~\cite{liu2011nonparametric},
matching~\cite{russell2009segmenting,tighe2010superparsing},
and distance evaluation~\cite{malisiewicz2008recognition})
and handcrafted features (\eg boundary~\cite{martin2004learning}
and superpixels~\cite{felzenszwalb2004efficient}).
Some unsupervised segmentation (US) methods
only focus on segmenting
objects but ignore the category,
while LUSS cares about object segmentation and classification.
Nevertheless, US models can provide prior knowledge to the LUSS model.
Numerous data-driven deep learning models have recently been developed
for supervised semantic segmentation
\cite{long2015fully,yu2015multi,badrinarayanan2017segnet,
	chen2017deeplab,Li_2020_CVPR}.
Based on pre-trained representations
\cite{ouali2020autoregressive,zhan2018mix,hwang2019segsort,van2020unsuperv},
a few unsupervised semantic segmentation (USS) models have been proposed
using segment sorting~\cite{hwang2019segsort},
mutual information maximization~\cite{ouali2020autoregressive},
region contrastive learning~\cite{van2020unsuperv},
and geometric consistency~\cite{cho2021picie}.
As the extension of USS,
LUSS differs from USS with its large-scale data and categories.
However, several issues limit the applicability of the USS to the LUSS task.
1) Existing methods focus on small datasets~
\cite{ouali2020autoregressive,zhan2018mix,hwang2019segsort,van2020unsuperv}
and a few (\eg 20+) easy categories~\cite{ouali2020autoregressive} (\eg sky and ground).
Because of the insufficient data,
the advantages of unsupervised learning of rich representations from large-scale data
are not explored.
The challenges of large-scale data
(\eg huge computational cost)
are also ignored.
2) Due to the lack of clear problem definition and standardized evaluation,
some methods utilize supervised prior knowledge, \eg
supervised pre-trained network weights~\cite{cho2021picie},
supervised edge detection~\cite{hwang2019segsort},
and supervised saliency detection
\cite{van2020unsuperv,GaoEccv20Sal100K,21PAMI-Sal100K},
making it difficult to evaluate these methods.

\subsection{Self-supervised Representation Learning}

The LUSS task relies on the semantic features provided by
self-supervised learning (SSL).
SSL approaches facilitate models learning semantic features with pretext tasks
\cite{jenni2018self,bojan17unsuperv,zhang2019aet,tsai2021selfsupervised},
\eg colorization~\cite{zhang2016colorful,iizuka2016let,larsson2017colorization},
jigsaw puzzles~\cite{noroozi2016unsupervised,noroozi2018boosting,misra2020self},
inpainting~\cite{pathak2016context},
adversarial learning~\cite{donahue2016adversarial,donahue2019large},
context prediction~\cite{doersch2015unsupervised,mundhenk2018improvements},
counting~\cite{noroozi2017representation},
rotation predictions~\cite{gidaris2018unsupervised,misra2020self},
cross-domain prediction~\cite{ren2018cross},
contrastive learning~\cite{oord2018representation,henaff2020data,He_2020_CVPR,chen2020simple,tian2020contrastive,tian2020makes},
non-contrastive learning
\cite{jean-bastien2020bootstrap,xie2020propagate,chen2020exploring},
and clustering
\cite{YM2020Self-labelling,li2020prototypical,caron2020unsupervised}.
We introduce several categories of SSL methods related to the LUSS task.

\myPara{Contrastive-based SSL.}
As the core of unsupervised contrastive learning methods
\cite{oord2018representation,hjelm2018learning,bachman2019learning,
	ye2019unsupervised,cao2020parametric,
	NEURIPS2020_63c3ddcc,wang2020unsupervised,patacchiola2020self,
	robinson2021contrastive},
instance discrimination with the contrastive loss~\cite{chopra2005learning,
	hadsell2006dimensionality,dosovitskiy2014discriminative}
considers images from different views~\cite{tian2020contrastive,tian2020makes}
or augmentations~\cite{chen2020simple,He_2020_CVPR} as pairs.
In addition, it forces the model to learn representations
by pushing ``negative'' pairs away and pulling ``positive'' pairs closer.
A memory bank~\cite{wu2018unsupervised} is introduced to enlarge
the available negative samples for contrastive learning.
MoCo~\cite{He_2020_CVPR} stabilizes the training with a momentum encoder.
CMC~\cite{tian2020contrastive} proposes contrastive learning from multi-views,
and SimCLR~\cite{chen2020simple} explores the effect of different
data augmentations.

\myPara{Non-contrastive-based SSL.}
Some non-contrastive approaches
\cite{xie2020propagate,tian2021understanding,zbontar2021barlow}
maximize the similarity of different types of outputs  of the image and
avoid negative pairs.
BYOL~\cite{jean-bastien2020bootstrap} predicts previous versions of
its outputs generated by the momentum encoder
to avoid outputs collapse to a constant trivial solution.
SimSiam~\cite{chen2020exploring} applies a stop-gradient operation
to avoid collapse.
Nevertheless, as the concept of the category is not included in
both contrastive and non-contrastive methods,
they are less effective for category-related tasks,
\ie instances from the same category not necessarily share
similar representations.

\myPara{Clustering-based SSL.}
Another line of work introduces a clustering strategy to unsupervised
learning~\cite{zhuang2019local,caron2018deep,caron2019unsupervised,
	ji2019invariant,yan2020clusterfit,Zhan_2020_CVPR,PCL}
that encourages a group of images to have feature representations
close to a cluster center.
Asano \etal~\cite{YM2020Self-labelling} propose simultaneous
clustering and representation learning by optimizing the same objective.
Li \etal~\cite{li2020prototypical}
maximize the log-likelihood of the observed data
via an expectation-maximization framework that
iteratively clusters prototypes and performs contrastive learning.
SwAV~\cite{caron2020unsupervised} simultaneously clusters views
while enforcing consistency between cluster assignments.
Compared to other representation learning methods,
the clustering strategy encourages stronger category-related representations
with category centroids.

\myPara{Pixel-level SSL.}
Some works uses self-supervised learning on the pixel-level
instead of image-level to enhance the transfer learning ability to
downstream tasks~\cite{xie2020propagate,roh2021spatially,wang2020DenseCL}.
PixPro~\cite{xie2020propagate} applies contrastive learning between
neighbour/other pixels and proposes pixel-to-propagation consistency
to enhance spatial smoothness.
SCRL~\cite{roh2021spatially} produces consistent spatial representations
of randomly cropped local regions with the matched location.
DenseCL~\cite{wang2020DenseCL} chooses positive pairs by
matching the most similar feature vectors in two views.
Despite the good performance for transfer learning,
these methods ignore the category-related representation ability
required by the LUSS task.

\subsection{Weakly Supervised Semantic Segmentation}

Weakly supervised semantic segmentation (WSSS)
\cite{wei2016stc,kolesnikov2016seed,li2018tell}
aims to carry out the task using weak annotations,
\eg image-level labels.
WSSS is related to LUSS as both require shape features.
However, some modules in typical WSSS methods,
\eg supervised ImageNet$_{1k}$ pre-trained models
\cite{ahn2019weakly,fan2020cian,sun2020mining,Chang_2020_CVPR},
image-level ground-truth labels~\cite{shimoda2019self,sun2020mining},
and large network architectures~\cite{Wang_2020_CVPR,Chang_2020_CVPR},
are not applicable to the LUSS tasks.
In addition, it is possible to use other alternative WSSS modules
\eg affinity prediction~\cite{ahn2018learning,ahn2019weakly,fan2020cian},
region separation~\cite{shimoda2019self,Fan_2020_CVPR},
boundary refinement~\cite{ahn2019weakly,chen2020weakly},
joint learning~\cite{zeng2019joint},
and sub-category exploration~\cite{Chang_2020_CVPR},
to improve LUSS models.

\begin{figure*}[t]
	\centering
	\addImg{1}{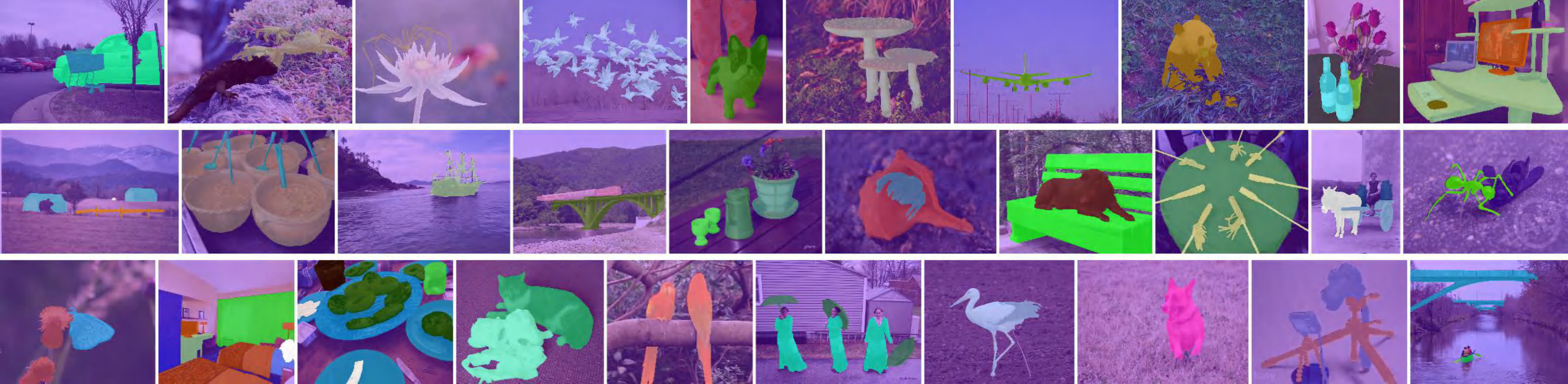}\\
	\vspace{-8pt}
	\caption{Visualization of the \ourdata~dataset.}
	\label{fig:vis_datasets}
\end{figure*}

\section{Large-scale Unsupervised Semantic Segmentation Benchmark}
\label{sec:luss}

The LUSS task aims to learn semantic segmentation from large-scale images
without direct/indirect human annotations.
Given a large set of images,
a LUSS model assigns self-learned labels to each pixel of all images.
For ease of understanding,
we give one of the possible pipelines for LUSS, as shown in
\secref{sec:basic_pipeline}.
A LUSS model simultaneously learns category and shape representations
from large-scale data without human annotation.
The model uses the learned feature representations for
label clustering and assignment to get the generated pixel-level labels.
Then, the model is fine-tuned on the generated labels to
refine the segmentation results.
Ideally, label assignment and refining
can be implicitly contained in the unsupervised representation learning process.

LUSS faces multiple challenges, \eg semantic representation learning,
category generation under large-scale data, and unsupervised setting.
Moreover, the lack of benchmarks limits the development of the LUSS task.
As such, we develop a LUSS benchmark with a clear objective,
large-scale training data, and comprehensive evaluation protocols.

\subsection{Large-scale LUSS Dataset: \ourdata}

The LUSS task is very challenging as it uses no human-annotated labels
for training and requires large-scale data to learn rich representations.
In principle, the scale of training images required by LUSS increases
with the growth of image complexity,
\ie large category numbers and complex senses require more training data.
Existing segmentation datasets can hardly support LUSS due to the large
image complexity but small data scale.
Some datasets,
\eg PASCAL VOC~\cite{Everingham15} and CityScapes~\cite{Cordts2016Cityscapes},
contain a limited number of images under a few scenes.
Other datasets, \eg ADE20K~\cite{zhou2018semantic},
COCO~\cite{lin2014microsoft}, and COCO-Stuff~\cite{caesar2018cvpr},
have complex images with a limited number of samples for each category,
which is hard for LUSS models to learn rich representations of complex senses
using limited data.

\begin{figure}[t]
	\centering \vspace{-.13in}
	\addImg{0.75}{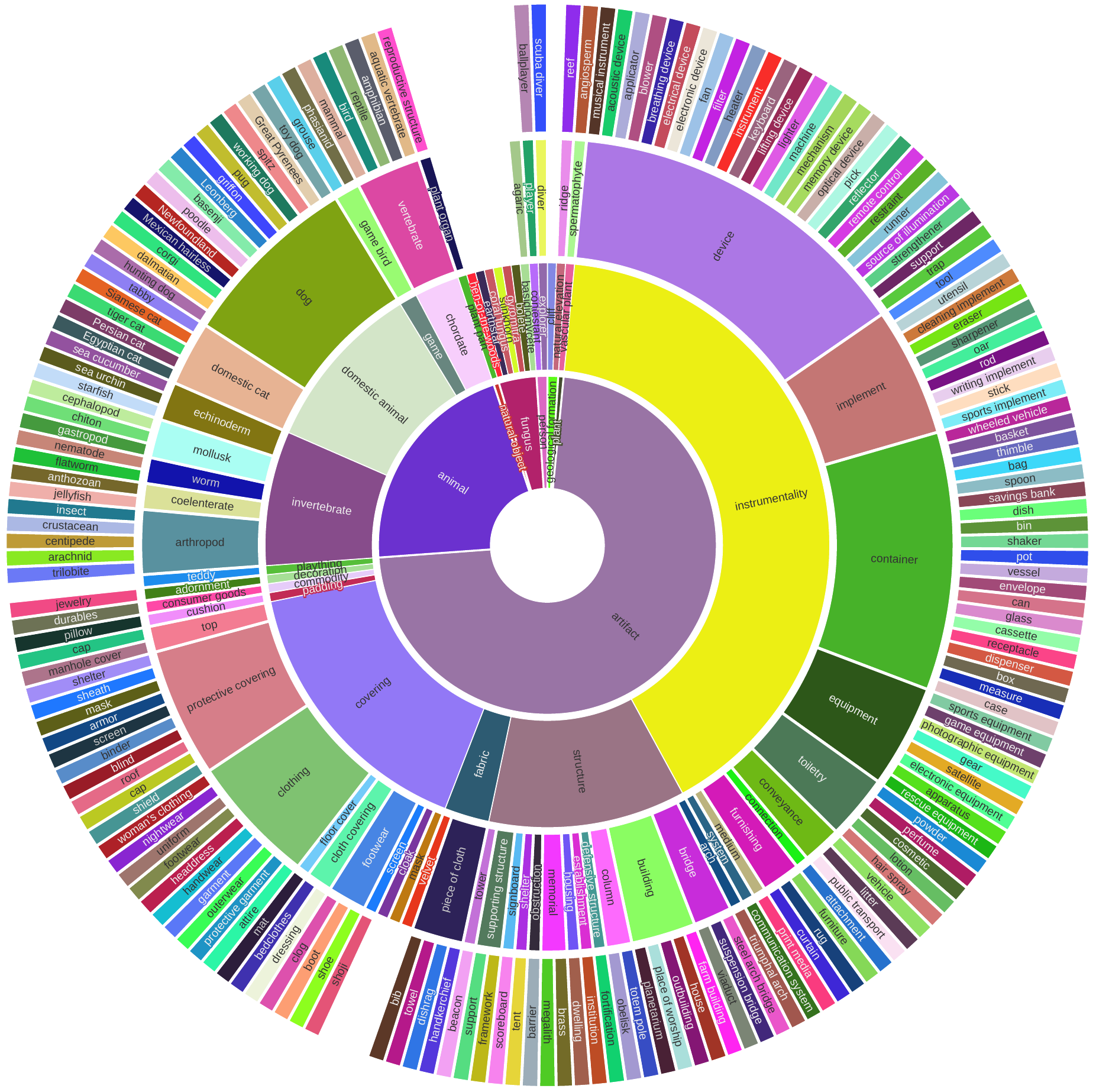}	\\
	\vspace{-8pt}
	\caption{Category structure tree of the \ourdata~dataset.}
	\label{fig:category_tree}
\end{figure}

To remedy drawbacks in these datasets, common supervised segmentation
approaches~\cite{chen2017deeplab,he2017mask,Chen_2018_ECCV}
fine-tune models pre-trained with
the widely used large-scale ImageNet dataset
\cite{He_2016_CVPR,xie2017aggregated,gao2019res2net,gao2021rbn}.
However, recent research~\cite{cole2021does,kotar2021contrasting} suggested
that performance on the ImageNet and downstream datasets is not
always consistent due to the inconstancy of data distribution,
data domain, and task objective.
For LUSS, fine-tuning pre-trained models on downstream datasets complicates the evaluation
and leads to possible unfair and biased comparisons.
ImageNet has diverse classes, a large data scale,
simple images, and sufficient images for each category,
making learning rich representations feasible.
Thus, ImageNet is widely used by most unsupervised learning methods
\cite{chen2020simple,caron2020unsupervised,xie2020propagate,tian2020contrastive,He_2020_CVPR}.
However, ImageNet has only image-level annotation,
and thus cannot be used for pixel-level evaluation of LUSS.
To facilitate the LUSS task,
we present a large-scale \ourdata~dataset by collecting data from the
ImageNet dataset~\cite{russakovsky2015imagenet}
and annotating pixel-level labels for LUSS evaluation.
We remove the unsegmentable categories, \eg bookshop,
and utilize \num{ 919 } categories in ImageNet.
As shown in~\tabref{tab:dataset_numbers},
the \ourdata~dataset (see \figref{fig:vis_datasets}) is much larger than
existing datasets in terms of image amount and category diversity
(see \figref{fig:category_tree}).

\subsubsection{Image Annotation.}

We annotate the validation/testing sets and parts of the training set
in the \ourdata~dataset for LUSS evaluation.
As the ImageNet dataset has incorrect labels and missing multiple categories,
we annotate the pixel-level semantic segmentation masks
following the relabeled image-level annotations in \cite{beyer2020we}
and further correct the missing and incorrect annotations.
The objects indicating the image-level labels are annotated,
and other parts are annotated as the `other' category.
`Other' means the categories are not frequently appeared in the dataset
or the surrounding environment.
For validation/testing sets,
we annotate all objects within the \num{ 919 } selected categories.
The parts that are difficult to distinguish are marked as `ignore',
which will not be used for evaluation.
For the training set, we randomly pick ten images for each category
and annotate objects corresponding to that category while
other objects belonging to the \num{ 919 } categories are labeled with `ignore'.

\def\PascalVOC{PASCAL VOC 2012~\cite{Everingham15}}
\def\CityScape{CityScapes~\cite{Cordts2016Cityscapes}}
\def\ADETWTK{ADE20K~\cite{zhou2018semantic}}

\begin{table}[t]
	\centering
	\setlength{\tabcolsep}{2.1mm}
	\caption{Categories and number of images comparison between
		the \ourdata~dataset and existing semantic segmentation datasets.
	}	\label{tab:dataset_numbers} \vspace{-7pt}
	\begin{tabular}{lcccc}  \toprule
		Dataset          & category    & train         & val         & test        \\ \midrule
		\PascalVOC       & \num{ 20 }  & \num{1464}    & \num{1449}  & \num{1456}  \\
		\CityScape       & \num{ 19 }  & \num{2975}    & \num{500}   & \num{1525}  \\
		\ADETWTK         & \num{ 150 } & \num{20210}   & \num{2000}  & \num{3000}  \\ \midrule
		\ourdata$_{50}$  & \num{ 50 }  & \num{64431}   & \num{752}   & \num{1682}  \\
		\ourdata$_{300}$ & \num{ 300 } & \num{384862}  & \num{4097}  & \num{9088}  \\
		\ourdata         & \num{ 919 } & \num{1183322} & \num{12419} & \num{27423} \\ \bottomrule
	\end{tabular}
\end{table}

\myPara{Semantic segmentation mask annotation.}
Given the categories of an image,
the annotator is asked to annotate the corresponding regions
and assign the correct categories.
The selected \num{ 919 } categories in the \ourdata~dataset have high diversity.
Some instances cannot be easily distinguished even with given categories,
\eg two breeds of dogs and uncommon things.
To reduce the difficulty for annotators to identify categories,
we splice four images with the same category into a four-square image.
In this case, annotators can easily distinguish the common categories
in the four-square image.
For images with multiple complex categories,
several groups of images containing the required categories are provided
to help annotators identify categories.
Since the categories in the \ourdata~dataset follow a tree-like structure
(see~\figref{fig:category_tree}),
different annotators are given images from different subsets of the
Word-Tree to further reduce the annotation difficulty.
Images with a resolution below \num{ 1000 } $\times$ \num{ 1000 } are resized to
\num{ 1000 } $\times$  \num{ 1000 }.
The annotator draws polygonal masks to the category-related regions with
about \num{ 400 } to \num{ 500 } points on the contour for each image.
Annotating on resized high-resolution images
results in precise pixel-level semantic segmentation masks.

\newcommand{\thCols}[1]{\multicolumn{3}{c}{#1}}

\begin{table}[t]
	\centering
	\setlength{\tabcolsep}{1.6mm}
	\caption{The number of categories in each image in the \ourdata~dataset.}
	\vspace{-7pt}
	\begin{tabular}{lccc|cccc}  \toprule
		                 & \multicolumn{6}{c}{Number of images}                                                                                \\ \cline{2-7}
		                 & \multicolumn{3}{c|}{val set}         & \thCols{test set}                                                            \\ \midrule
		Categories in each image
		                 & \num{ 1 }                            & \num{ 2 }         & $>$2        & \num{ 1 }     & \num{ 2 }    & $>$2        \\ \midrule
		\ourdata$_{50}$  & \num{ 745 }                          & \num{ 7 }         & \num{ 0 }   & \num{ 1676 }  & \num{ 6 }    & \num{ 0 }   \\
		\ourdata$_{300}$ & \num{ 3971 }                         & \num{ 118 }       & \num{ 8 }   & \num{ 8815 }  & \num{ 264 }  & \num{ 9 }   \\
		\ourdata         & \num{ 11294 }                        & \num{ 954 }       & \num{ 171 } & \num{ 25133 } & \num{ 1938 } & \num{ 352 } \\ \bottomrule
	\end{tabular}
	\label{tab:num_cat}
\end{table}

\begin{figure}[t]
	\centering
	\scriptsize
	\hfill
	\subfigure[Number of images per class.]{
		\begin{overpic}[width=0.47\linewidth]{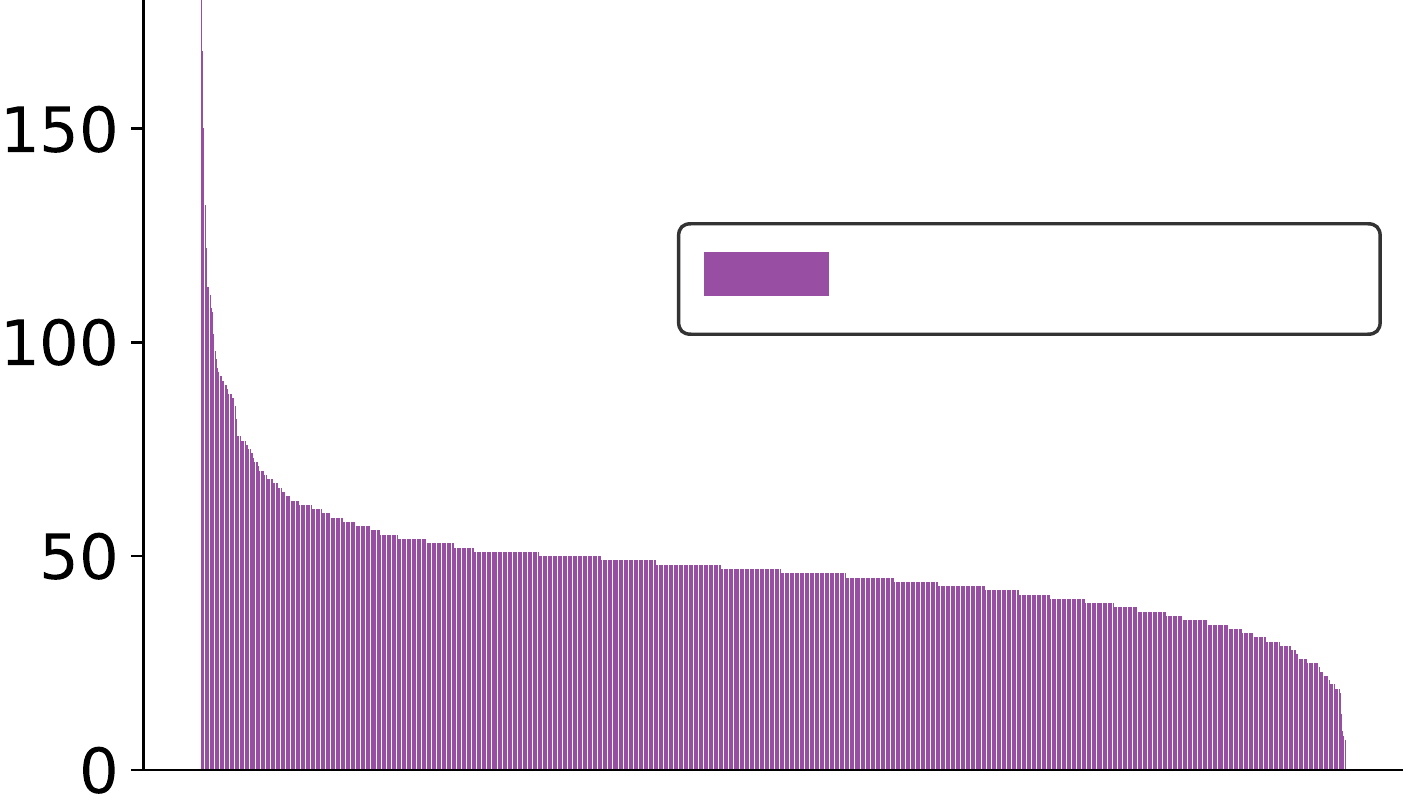}
			\put(61, 36.5){val and test set}
			\put(-8, 16){\rotatebox{90}{number}}
		\end{overpic}
	}
	\subfigure[Number of images per class.]{
		\begin{overpic}[width=0.47\linewidth]{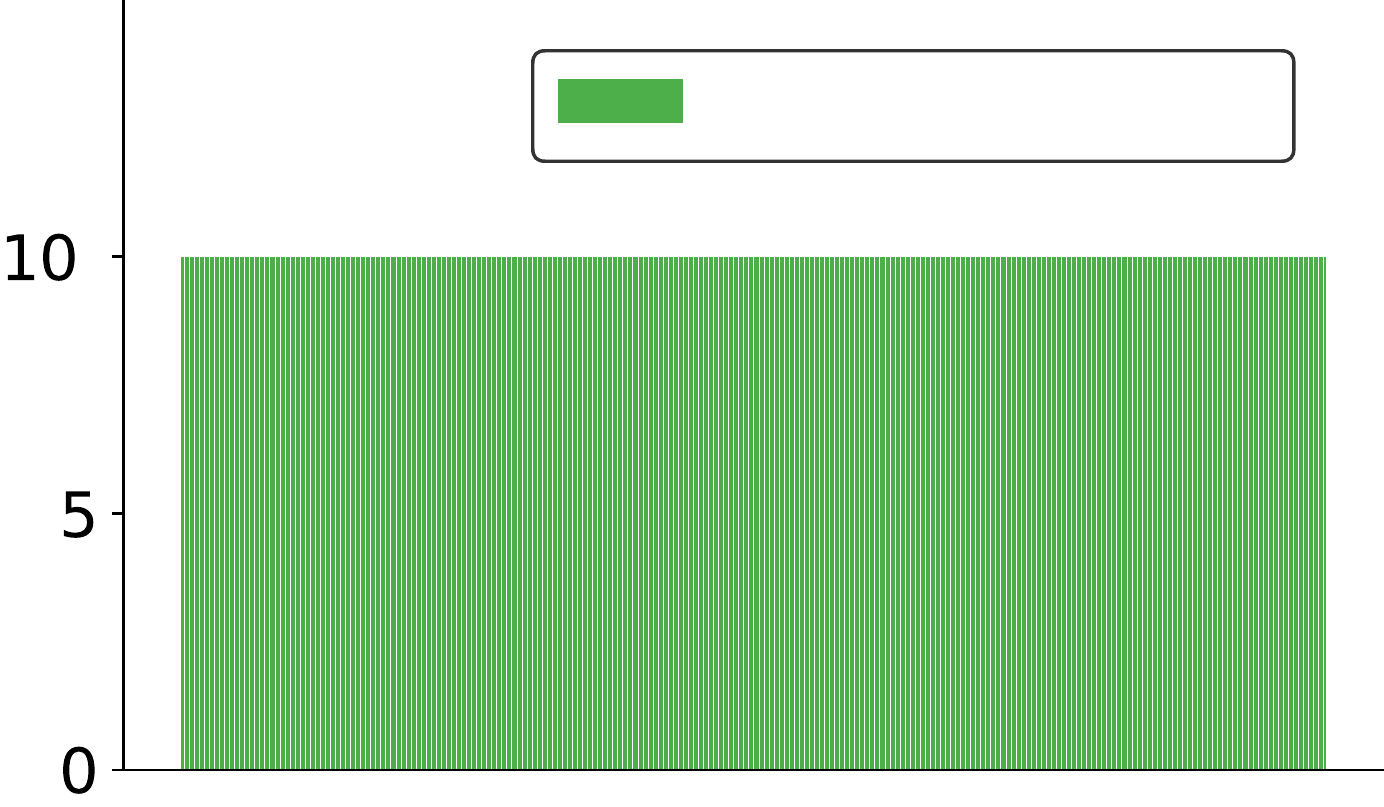}
			\put(52, 49.6){anno-training set}
		\end{overpic}
	} \\ \hfill
	\subfigure[Number of pixels per class.]{
		\begin{overpic}[width=0.47\linewidth]{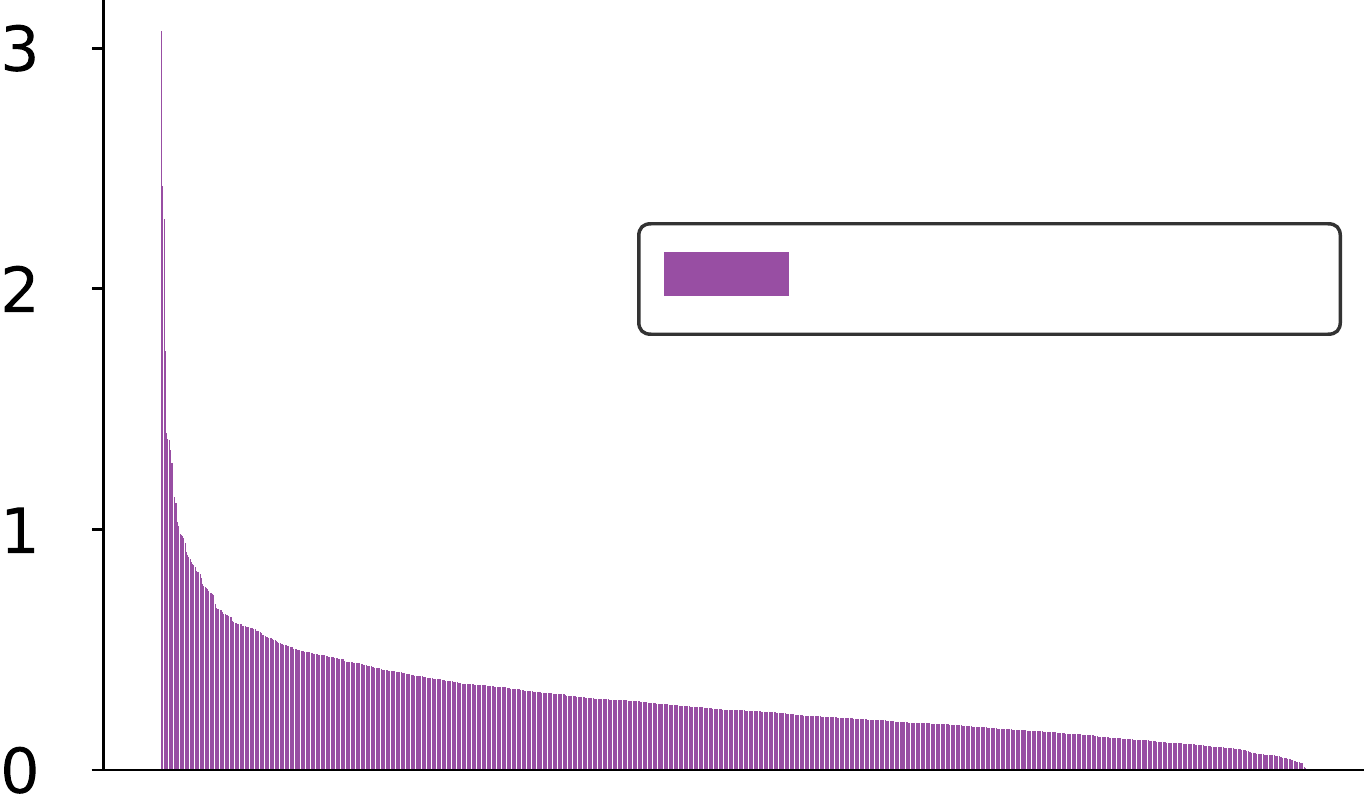}
			\put(60.5, 37.6){val and test set}
			\put(-8, 16){\rotatebox{90}{number}}
			\put(10, 54){$\times 10^{7}$}
		\end{overpic}
	}
	\subfigure[Number of pixels per class.]{
		\begin{overpic}[width=0.47\linewidth]{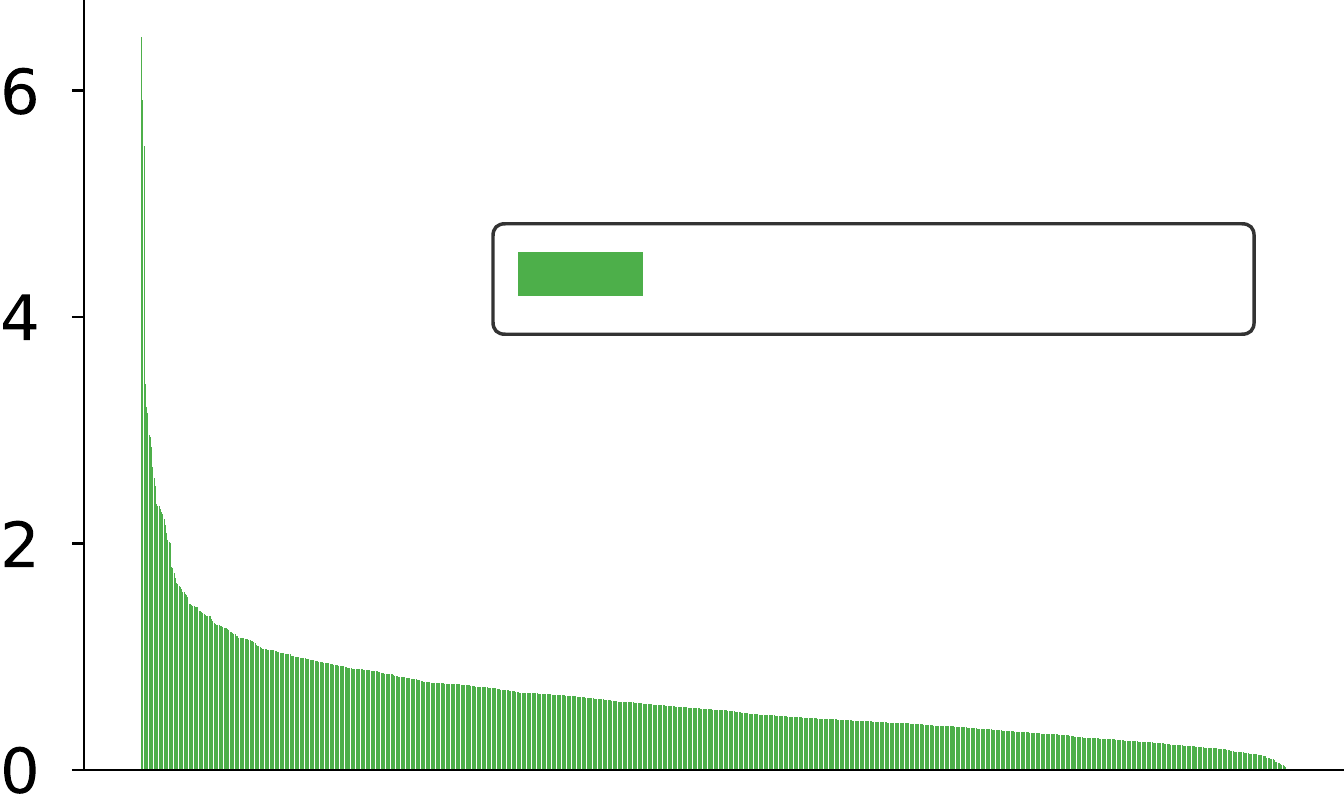}
			\put(51., 38.2){anno-training set}
			\put(9, 52){$\times 10^{6}$}
		\end{overpic}
	}\\ \vspace{-6pt}
	\caption{Instance-level/pixel-level number distribution
		among categories of the \ourdata~dataset,
		\ie the number of images/pixels per class.
	}\label{fig:category_distributation_datasets}
\end{figure}
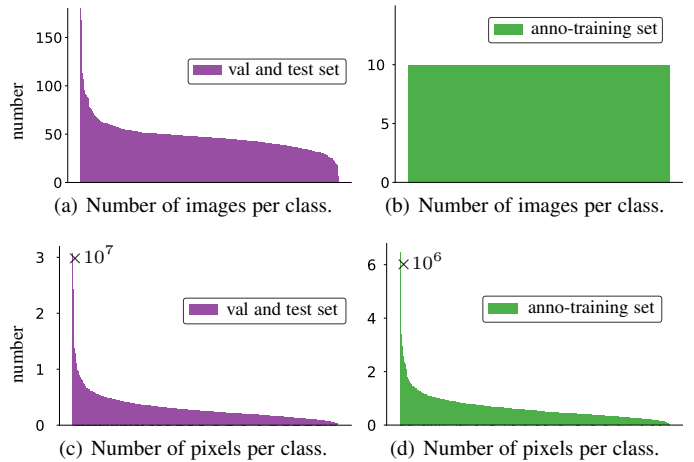

\myPara{Labeling process.}
The annotation team for this dataset contains an organizer,
four quality inspectors, and \num{ 15 } annotators.
We introduce the labeling process as follows:

{\noindent \bf Step 1.}  Annotators were instructed on how to annotate labels.
Then annotators were asked to annotate a group of randomly picked images
following instructions.
Quality inspectors checked these annotated images,
and failure cases were corrected and shown as examples to all annotators.

	{\noindent \bf Step 2.}
The annotators were divided into several groups,
and each group has a group leader.
The organizer then assigned images to each group of annotators.
After annotating the images,
the group leader summarized all annotations and checks the annotation quality.
Other annotators checked the annotations from the group leader in the group.

	{\noindent \bf Step 3.}
The checked annotations were then given to the quality inspectors.
Quality inspectors checked the annotations and give feedback regarding
the failure cases.
Common failure cases and corresponding explanations are
sent to all groups to improve the annotation quality of the following images.

	{\noindent \bf Step 4.}
The organizer then sampled images and checked the corresponding annotations
to ensure the annotation quality.

\myPara{Correct missing/incorrect labels.}
During the labeling process,
we observed that there were still some missing and incorrect
image-level annotations in~\cite{beyer2020we} due to the high diversity
and large-scale properties of ImageNet.
Therefore, we presented several schemes to correct labels as much as possible:
1) We found that some categories are related to each other,
\eg the spider and spider web usually appear in the same image.
Based on the initial human-observed missing categories,
we double-checked images whose categories are related to other categories.
2) We used supervised image-level classifiers,
\eg Swin transformer~\cite{Liu_2021_ICCV} and Res2Net~\cite{gao2019res2net},
to help find missing categories by checking the labels predicted with
high confidence but not in the ground-truth.
With these schemes, we managed to correct \num{ 296 } mislabeled images and find \num{ 942 } images with missing labels.

\subsubsection{Statistics and distribution.}
\label{sec:dataset_state}

\myPara{Image numbers.}
As shown in~\tabref{tab:dataset_numbers},
after removing the unsegmentable categories in the ImageNet dataset,
\eg bookshop, valley, and library,
the \ourdata~dataset contains \num{ 1183322 } training, \num{ 12419 } validation,
and \num{ 27423 } testing images from \num{ 919 } categories.
Many existing self-supervised representation learning methods
\cite{He_2020_CVPR,xie2020propagate}
are trained with the ImageNet dataset.
For a fair comparison,
we use the full ImageNet dataset that contains \num{ 1281167 } training images
for unsupervised representation learning and
utilize the \ourdata~training set for other processes in LUSS.
We annotate \num{ 39842 } validation/testing images and \num{ 9190 } training images
with precise pixel-level masks,
and some visualized annotations are shown in~\figref{fig:vis_datasets}.
Our pixel-level labeling enables the \ourdata~dataset with multiple labels
in each image.
\tabref{tab:num_cat} gives the number of categories per image in
the \ourdata~validation/testing sets.
A majority of images contain one category,
and 8.6\% of images have more than one category.
\ourdata~has simpler images and more categories than existing
semantic segmentation datasets,
which is suitable for the LUSS task considering the difficulty caused
by no human annotation and large image and category numbers.

\myPara{Category distribution.}
As shown in~\figref{fig:category_tree},
categories in the \ourdata~dataset show a tree-like structure as
they are extracted from the Word-Tree~\cite{russakovsky2015imagenet}.
\figref{fig:category_distributation_datasets} shows
the image-level and pixel-level number distribution among categories of the
\ourdata~dataset, \ie the number of images/pixels per class.
The training set and validation/testing sets have similar distributions.
The number of images for most categories is balanced,
while the number of pixels per category presents the long-tail distribution.
The imbalanced pixel-level category distribution may introduce new challenges
that are not considered in the image-level representation learning.
Compared to the original ImageNet dataset with a similar number of images
for each category in the validation set,
the relabeled \ourdata~validation/testing sets presents a more
unbalanced number of images over categories.

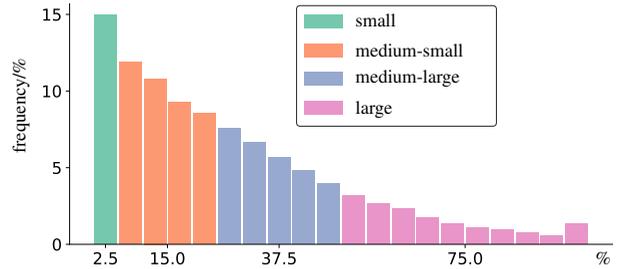
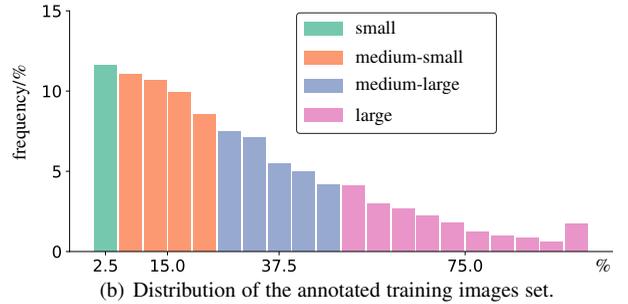
\begin{figure}[t]
	\centering
	\scriptsize
	\subfigure[Distribution of validation and testing set.]{
		\begin{overpic}[width=.85\linewidth]{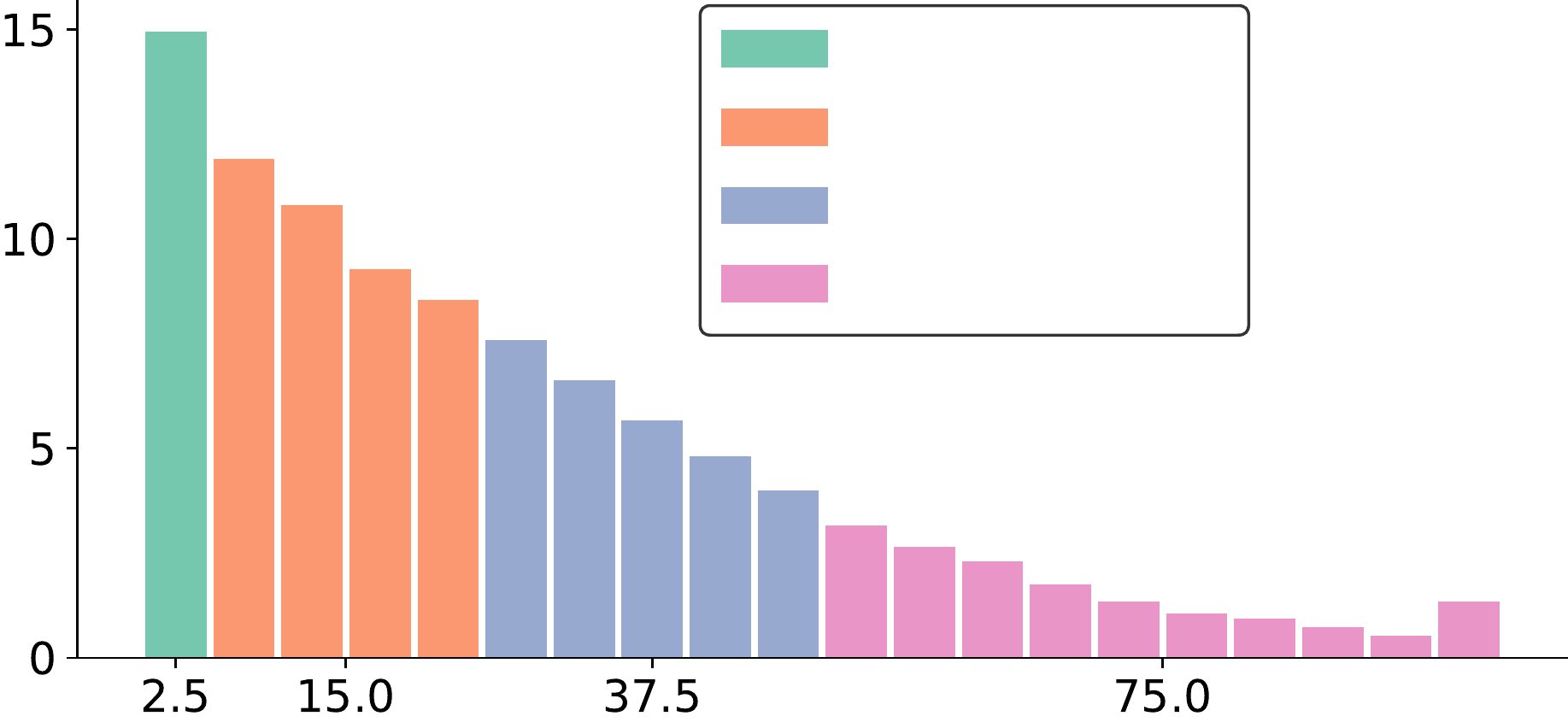}
			\put(55, 42){small}
			\put(55, 36.9){medium-small}
			\put(55, 32.2){medium-large}
			\put(55, 26.8){large}
			\put(97, 0.6){\%}
			\put(-5, 20){\rotatebox{90}{frequency/\%}}
		\end{overpic}
	}
	\subfigure[Distribution of the annotated training images set.]{
		\begin{overpic}[width=.85\linewidth]{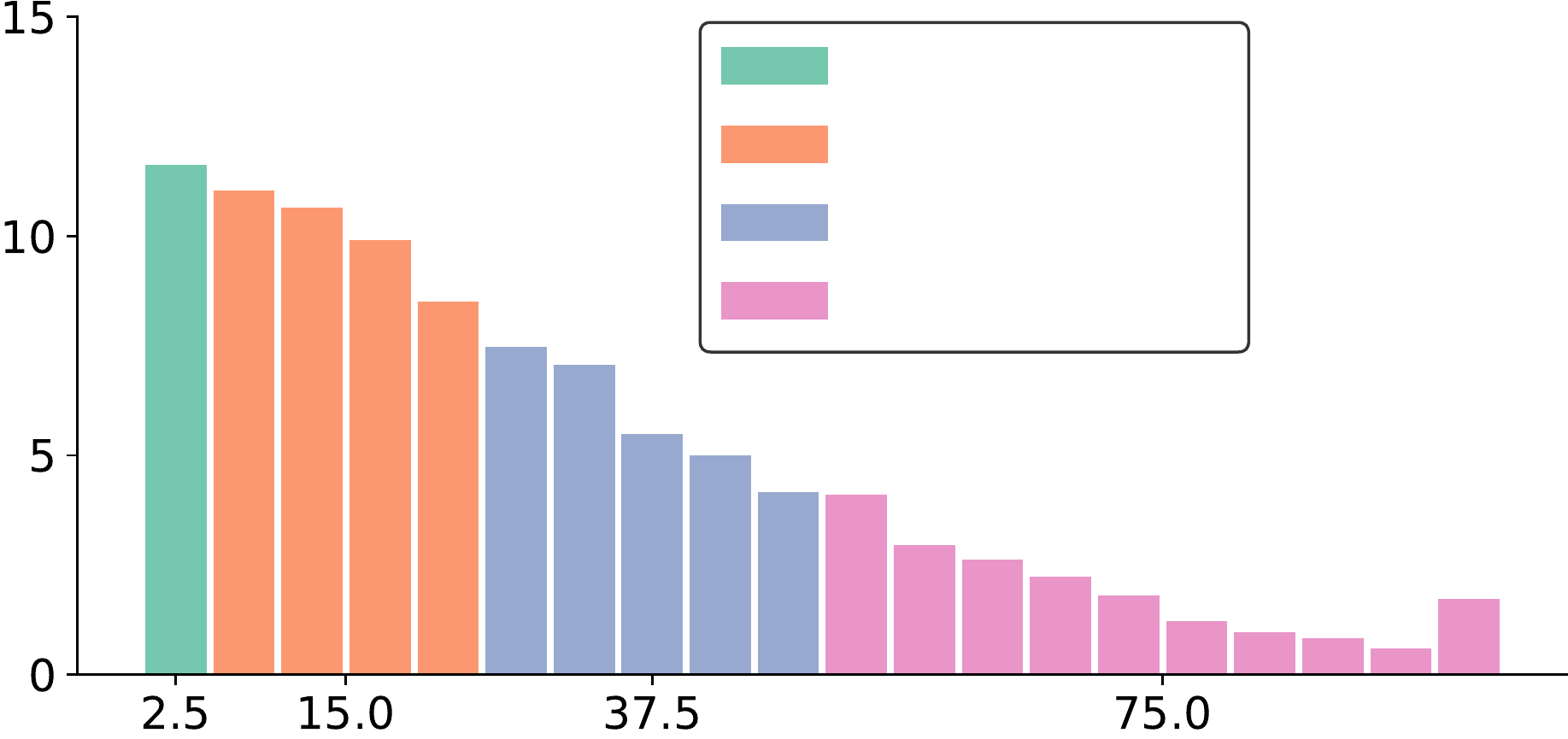}
			\put(55, 42){small}
			\put(55, 36.9){medium-small}
			\put(55, 32.2){medium-large}
			\put(55, 26.8){large}
			\put(97, 0.6){\%}
			\put(-5, 20){\rotatebox{90}{frequency/\%}}
		\end{overpic}
	}\vspace{-6pt}
	\caption{Distribution of object size in the \ourdata~dataset.
		The object size is defined as the ratio of object size to the image size.
	}\label{fig:visualization_pbject_size_datasets}
\end{figure}

\myPara{Object size.}
As it is more difficult to segment smaller objects,
we divide objects into groups,
\ie small (0\%-5\%), medium-small (5\%-25\%),
medium-large (25\%-50\%), and large object size (50\%-100\%),
according to the ratio of object size to the image size.
The object size distribution shown in
\figref{fig:visualization_pbject_size_datasets}
indicates that most objects are relatively small.

\myPara{Position distribution.}
We superimpose segmentation masks from the validation and testing sets
to analyze the position distribution of semantic objects in the dataset,
as shown in~\figref{fig:position_distribution_datasets} (top).
The objects in \ourdata~tend
to be in the centre of the image,
which explains the effectiveness of the central crops strategy of
existing self-supervised methods~\cite{chen2020simple,He_2020_CVPR}.
We also superimpose the boundary of objects,
as shown in~\figref{fig:position_distribution_datasets} (down).
It shows that objects cover almost all areas instead of only the
central area of images.
In addition, we compare the distributions of our dataset
with COCO~\cite{lin2014microsoft} and Open Images~\cite{OpenImages} datasets in~\figref{fig:position_distribution_datasets}.
Our dataset and the other two datasets have similar distributions.
The centre-skewed distribution is observed for all datasets,
and we assume that humans might tend to record more centre-biased images.
Interestingly, the distribution map of ImageNet-S is almost identical to
the Open Images dataset, famous for its real-life property.

\myPara{Annotation consistency.}
We validate the annotation quality by evaluating
the annotation consistency of different people.
We have asked four annotators to annotate the same 100 randomly picked images, respectively.
Based on four sets of samples,
we evaluate the average metrics between each pair
using mask mIoU and boundary mIoU, as shown in~\tabref{tab:consistency}. %
The mask mIoU is 98.7\%, showing a high annotation consistency.
With a small d of 2\%, the boundary mIoU still has 92.4\%,
indicating high constant boundary annotations.
We visually observe that the main annotation
differences are in the boundary regions.
By comparing objects of different
sizes, smaller objects have lower
annotation consistency
since the boundary regions occupy
a larger proportion of the annotation mask in smaller objects.

\newcommand{\fRows}[1]{\multirow{3}*{#1}}

\begin{table}[t]
	\caption{The consistency among four annotators using \num{ 100 } randomly picked images.
		The $d$ indicates the pixel distance as in \cite{cheng2021boundary}.
	}
	\centering
	\footnotesize
	\setlength{\tabcolsep}{1.8mm}
	\begin{tabular}{lccccccccccccccc}
		\toprule
		Metrics   & $d$ & All  & S.   & M.S. & M.L. & L.   \\ \midrule
		\fRows{Boundary mIoU\cite{cheng2021boundary}}
		          & 2\% & 92.4 & 91.0 & 91.5 & 92.7 & 93.4 \\
		          & 3\% & 94.8 & 92.6 & 93.9 & 95.1 & 95.5 \\
		          & 4\% & 95.9 & 93.2 & 95.1 & 96.2 & 96.5 \\ \midrule
		Mask mIoU & -   & 98.7 & 93.4 & 97.1 & 99.0 & 99.3 \\
		\bottomrule
	\end{tabular}
	\label{tab:consistency}
\end{table}

\renewcommand{\addImg}[2]{\includegraphics[width=#1\linewidth]{figures/datasets/#2}}
\begin{figure}[t]
	\centering
	\renewcommand{\arraystretch}{0.2}
	\setlength{\tabcolsep}{0.2mm}
	\begin{tabular}{cccccccc}
		\addImg{0.3}{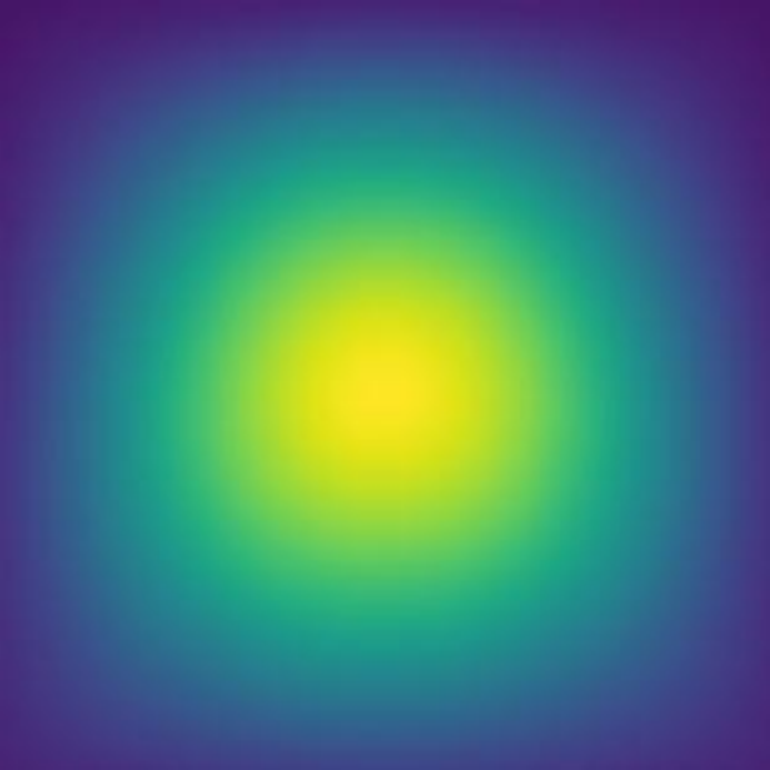}           &
		\addImg{0.3}{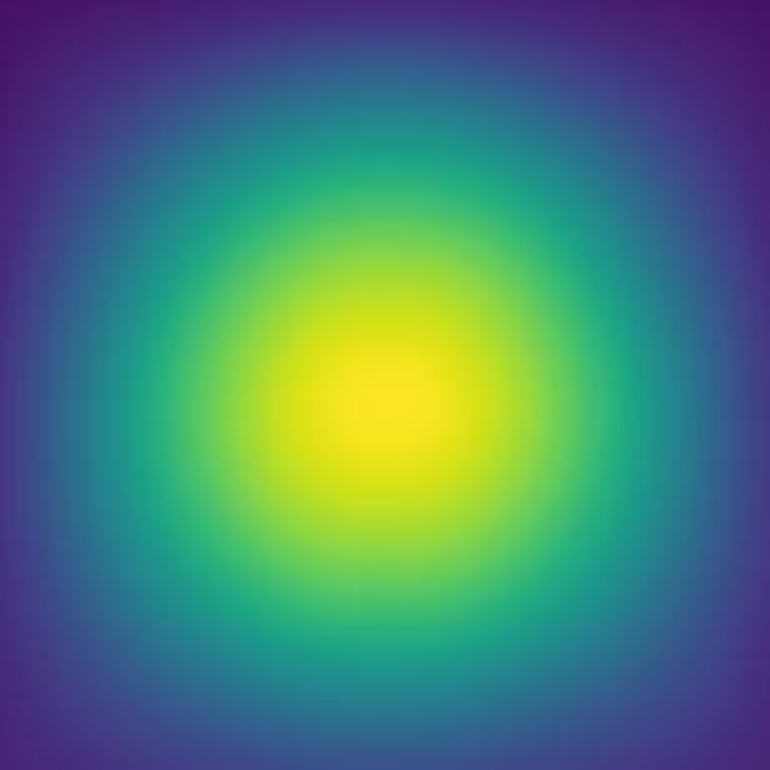} &
		\addImg{0.3}{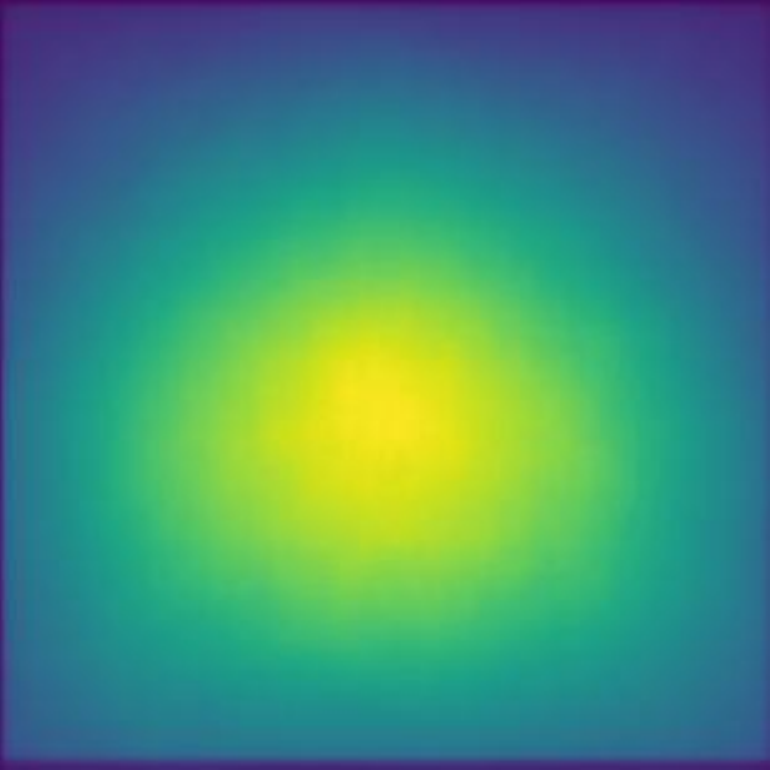}      &                    \\
		\addImg{0.3}{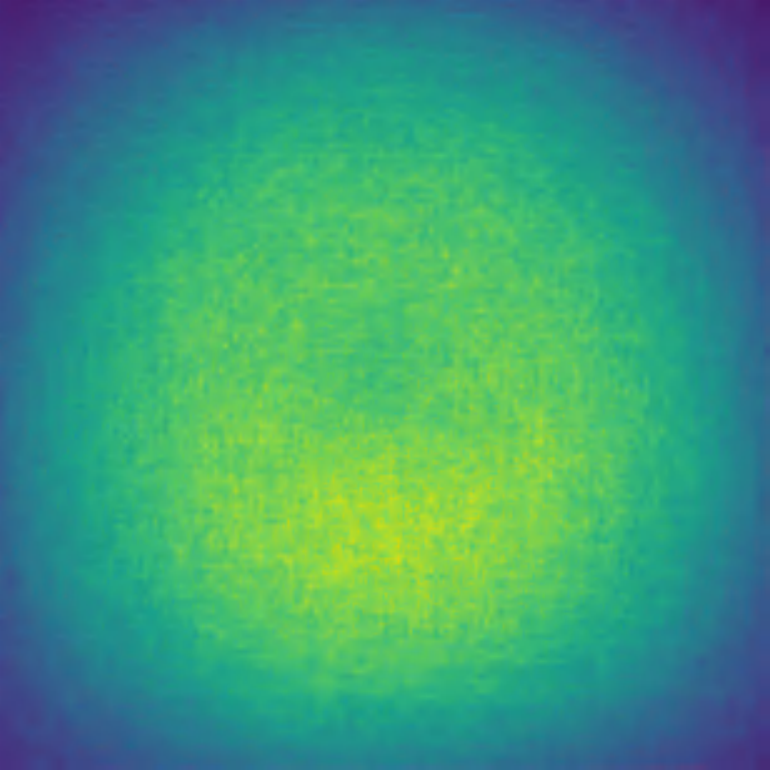}                 &
		\addImg{0.3}{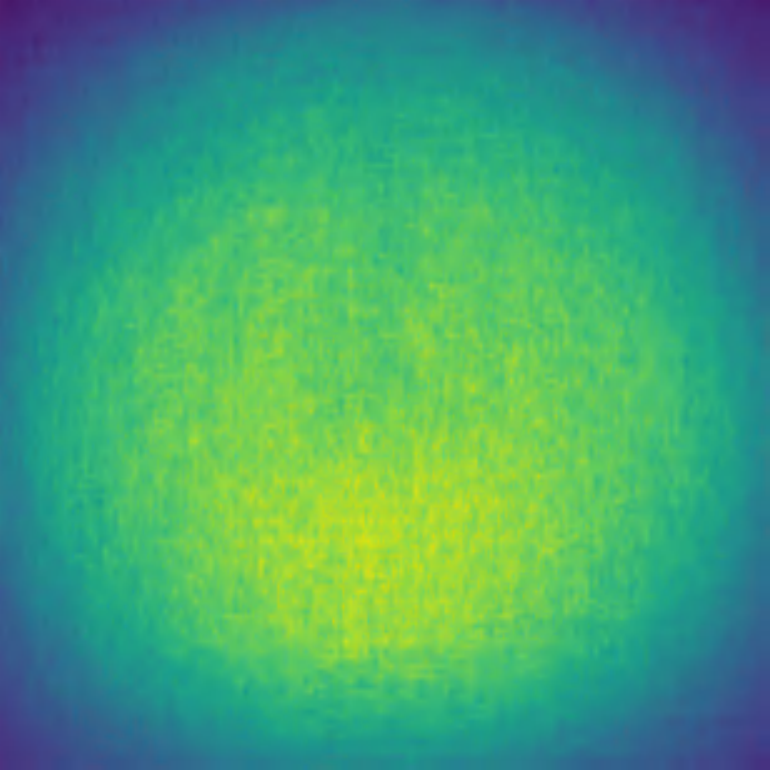}       &
		\addImg{0.3}{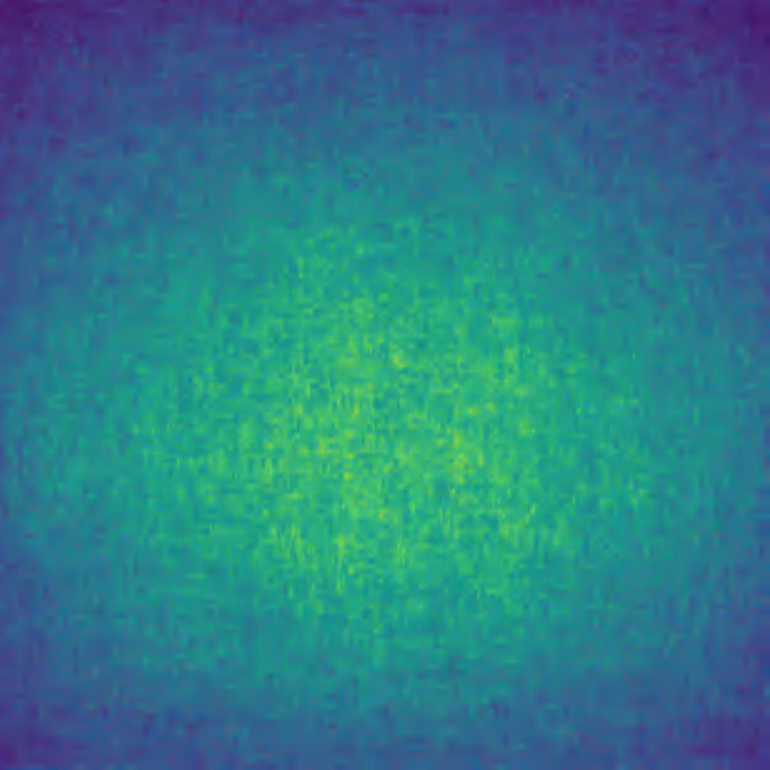}                                 \\
		ImageNet-S                             & Open Images & COCO \\
	\end{tabular}
	\caption{Position distribution comparison among datasets:
		(top) the position distribution for segmentation masks,
		(down) the position distribution for mask boundaries.
	}
	\label{fig:position_distribution_datasets}
\end{figure}

\begin{figure*}[t]
	\centering
	\begin{overpic}[width=0.9\linewidth]{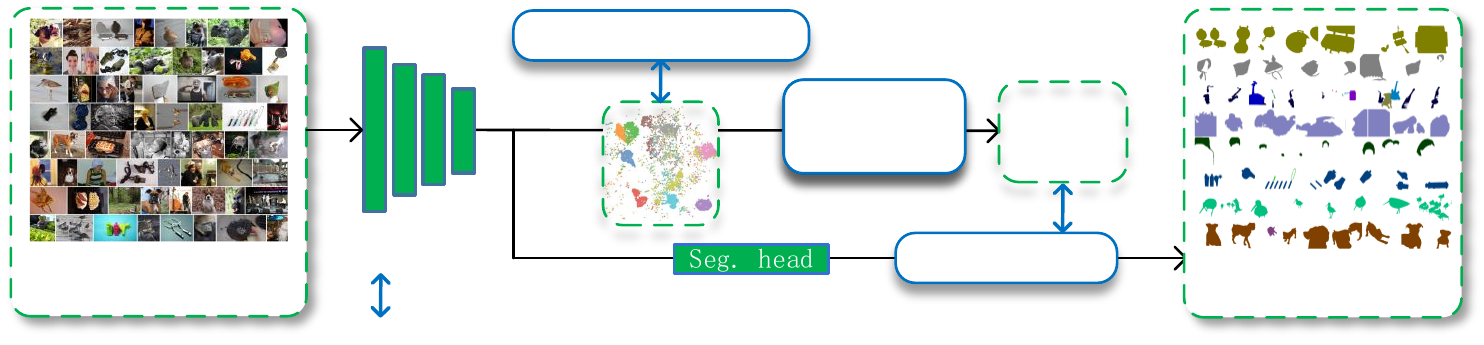}
		\put(3.5,4){Large-scale data}
		\put(27,2.5){Supervision}
		\put(37.5,20.2){Self-supervision}
		\put(54.5,14){Clustering}
		\put(68.7,15){Pseudo}
		\put(68.7,12.5){Labels}
		\put(63.0,5.2){Fine-tuning}
		\put(83.3,3.4){Semantic seg.}
	\end{overpic}\\
	\vspace{-8pt}
	\caption{The pipeline of our method for the LUSS task.
	}\label{fig:pipeline_LUSS}
\end{figure*}

\myPara{\ourdata-50/300 under a limited budget.}
To facilitate the research under a low computational budget,
we develop two subsets containing \num{ 50 } and \num{ 300 } categories,
namely \ourdata$_{50}$ and \ourdata$_{300}$.
Considering the difficulty of the LUSS task,
we choose \num{ 50 } distinguishable categories in daily life for \ourdata$_{50}$.
The \ourdata$_{300}$ is composed of \ourdata$_{50}$ and \num{ 250 } randomly
sampled categories.
The number of images in \ourdata$_{50}$ and \ourdata$_{300}$ are shown
in \tabref{tab:dataset_numbers}.
Even the \ourdata$_{50}$ subset has more images than
most semantic segmentation datasets.

\subsection{Evaluation}

\subsubsection{Evaluation Protocols}
\label{sec:protocols}

Due to the lack of ground-truth (GT) labels during training,
LUSS models cannot be directly evaluated as in the supervised setting.
We present three evaluation protocols for LUSS,
including the fully unsupervised evaluation,
semi-supervised evaluation, and distance matching evaluation.

\myPara{Fully unsupervised protocol.}
The fully unsupervised evaluation protocol requires no human-annotated labels
during training and only needs the validation/testing set for evaluation.
Unlike the supervised tasks,
categories are generated by the model in the LUSS task,
which needs to be matched with GT categories during evaluation.
We present the default image-level matching scheme,
while a more effective matching scheme should improve LUSS evaluation performance.
Suppose the set for matching (normally validation set) has
$N$ images and $C$ categories.
The number of categories is implicitly contained in the training dataset as the dataset has $C$ major categories.
We assume the unsupervised model should learn to
generate more than $C$ categories from the dataset during training.
The default image-level matching scheme only matches $C$ generated categories with $C$ ground-truth categories.
Given the image set $\mathbf{D}=\{\mathbf{D}_k, k \in [1,N] \}$
with GT labels $\mathbf{G}=\{ \mathbf{G}_k, k \in [1,N]\}$
and predicted labels $\mathbf{P}=\{\mathbf{P}_k, k \in [1,N] \}$,
where $\mathbf{G}_k$ and $\mathbf{P}_k$ are the GT and
predicted category sets of the image $\mathbf{D}_k$.
We calculate the matching matrix $\mathbf{S}\in \Real^{C\times C}$
between generated and GT categories,
in which $\mathbf{S}_{ij}$,
representing the matching degree between the $i$-th generated category
and the $j$-th GT category,
is larger when two categories are more likely to be the same category:
\begin{equation}
	\mathbf{S}_{ij} = \sum_{k=1}^{N} \mathbb I \{(i,j) \in
	\mathbf{P}_{k} \times \mathbf{G}_{k} \},
\end{equation}
where $\mathbf{P}_{k} \times \mathbf{G}_{k}$ is the Cartesian product
of $\mathbf{P}_{k}$ and $\mathbf{G}_{k}$,
and the indicator $\mathbb I$ equals \num{ 1 } when $(i,j)$ belongs to
$\mathbf{P}_{k} \times \mathbf{G}_{k}$.
With the matching matrix $\mathbf{S}\in \Real^{C\times C}$,
we find the bijection $\mathbf{f}: i \mapsto j$ between generated
and GT categories
using the Hungarian algorithm~\cite{kuhn1955hungarian} to maximize $\sum_{i=1}^{C}\mathbf{S}_{i, \mathbf{f}(i)}$.
We observe that there are failed matching cases and
some generated categories are not in the ground-truth categories,
indicating the limit of our baseline matching method.
We expect future works to propose more effective matching methods to solve this problem.

\myPara{Semi-supervised protocol.}
We can conduct semi-supervised fine-tuning to evaluate LUSS models
as we annotate about 1\% of training images with pixel-level labels.
The semi-supervised evaluation protocol requires fine-tuning the trained
LUSS models with human-labeled training images.
Therefore, this protocol does not need matching generated and GT category.
Also, this protocol is suitable for real-world applications where
a small part of images are labeled and many images are unlabeled.

\myPara{Distance matching protocol.}
In the distance matching evaluation protocol,
we directly get the embeddings of GT categories with
the pixel-level labeled training images
and match them with embeddings in validation/testing sets to assign labels.
Specifically, for pixels with the same category in an image
(including the `other' category),
we get the averaged embeddings and the corresponding label in the training set.
Then we infer the segmentation masks of the validation/testing sets
using the k-NN classifier~\cite{wu2018unsupervised}.
For each pixel embedding in the validation/testing sets,
we find the top-$k$ similar embeddings in the training set and
the corresponding labels.
The assigned label of each pixel is determined by the voting among these
$k$ labels.

\subsubsection{Evaluation metrics}
\label{metric}
We use the mean intersection over union (mIoU), boundary mIoU (b-mIoU), image-level accuracy (Img-Acc),
and F-measure ($F_{\beta}$) as the evaluation metrics for the LUSS task.
During the evaluation,
all images are evaluated with the original image resolution.
mIoU and b-mIoU are comprehensive evaluation metrics,
while Img-Acc and $F_{\beta}$ explain the performance
from category and shape aspects.
We give the implementation details of these evaluation metrics
in the supplementary.

\myPara{Mean IOU.}
Similar to the supervised semantic segmentation task
\cite{Everingham15,zhou2018semantic},
we utilize the mIoU metric
to evaluate the segmentation mask quality.
Apart from the major categories, the `other' category
is also considered to get mIoU.

\myPara{Boundary mean IoU.}
Unlike the mask mIoU above that measures all object regions,
the boundary mean IoU (b-mIoU)~\cite{cheng2021boundary} focuses on the boundary
regions.
We use the boundary mIoU to measure the semantic segmentation quality
of boundary regions.
According to the segmentation consistency analysis in~\secref{sec:dataset_state},
we use $d=3\%$ for boundary IoU~\cite{cheng2021boundary}.

\myPara{Image-level accuracy.}
The Img-Acc can evaluate the category representation ability of models.
As many images contain multiple labels,
we follow~\cite{beyer2020we} and treat the predicted label as correct
if the predicted category with the largest area is within the GT label list.

\myPara{F-measure.}
In addition to category-related representation,
we utilize $F_{\beta}$ to evaluate the shape quality \cite{ChengPAMI},
which ignores the semantic categories.
We treat major categories as the foreground category and
the `other' category as the background category.

\section{A LUSS Method}
\label{sec:basic_pipeline}

\subsection{Overview}

We summarize the main challenges of the LUSS task:
1) The model should learn category-related representations
without image-level label supervision.
2) Extracting a semantic segmentation mask
requires the model to learn shape representations.
3) Shape and category representations should coexist with minimal conflict.
4)~With learned representations,
the model should assign self-learned labels to each pixel in the image
with high efficiency.
5)~The large-scale training data helps to learn rich representations
in an unsupervised learning manner
but inevitably causes a large amount of training cost,
which requires improving the training efficiency.

Considering the above challenges,
we propose a new method for LUSS, namely PASS, (see~\figref{fig:pipeline_LUSS}),
containing four steps.
1)
A randomly initialized model is trained with {\bf self-supervision}
of pretext tasks to learn shape and category representations.
After representation learning,
we obtain the features set for all training images.
2)
We then apply a pixel-attention-based {\bf clustering} scheme to obtain
pseudo categories and assign generated categories to each image pixel.
3)
We {\bf fine-tune} the pre-trained model with the generated pseudo labels to
improve the segmentation quality.
4)
During {\bf inference}, the LUSS model assigns generated labels to
each pixel of images, same to the supervised model.
Noted that the pipeline in our method is not the only option,
and other pipelines are also encouraged for the LUSS task.
We now give a detailed introduction of each step as follows.
Some frequently used symbols are listed in \tabref{tab:symbols} for
easier understanding.

\subsection{Unsupervised Representation Learning}

For the first step in our LUSS method,
a randomly initialized model, \eg ResNet,
is trained with self-supervision of pretext tasks to learn semantic
representations.
The LUSS task requires category-related representation to
distinguish scenes from different classes and
shape-related representation to form the shape of objects.
Prior works have made many efforts to learn image-level
category-related representation or pixel-level representation
\cite{xie2020propagate,roh2021spatially,wang2020DenseCL}.
However, the image-level methods often ignore shape-related features.
The pixel-level methods focus on the transfer learning performance on
supervised downstream tasks.
As observed by~\cite{zhao2021what},
the performance of most downstream tasks relies on low-level feature
from the early stage of the network.
Thus, pixel-level methods that perform well on downstream tasks
may not learn high-level semantic features with category and shape information.

To obtain a powerful representation for LUSS,
we present two self-supervised learning strategies to enhance both category
and shape representation, including
1) a non-contrastive pixel-to-pixel representation alignment strategy
to enhance the pixel-level shape-related representation without
hurting the instance-level category representation.
2) a deep-to-shallow supervision strategy to enhance the
representation quality of mid-level features of the network.

\def\StopG{\cancel{\mathbf{G}}}
\def\tdV{\mathbf{\tilde{v}}}
\def\tdZ{\mathbf{\tilde{z}}}
\def\sRes{\times H \times W}
\def\feaZ{\mathbf{Z}}
\def\feaQ{\mathbf{Q}}
\def\feaY{\mathbf{Y}}
\def\featz{\mathbf{z}}
\def\featq{\mathbf{q}}
\def\featy{\mathbf{y}}
\def\mlp{\mathbf{M}}
\def\mProj{\mathbf{M_p(\tilde{z})}}

\begin{figure}[t]
	\centering
	\small
	\begin{overpic}[width=\linewidth]{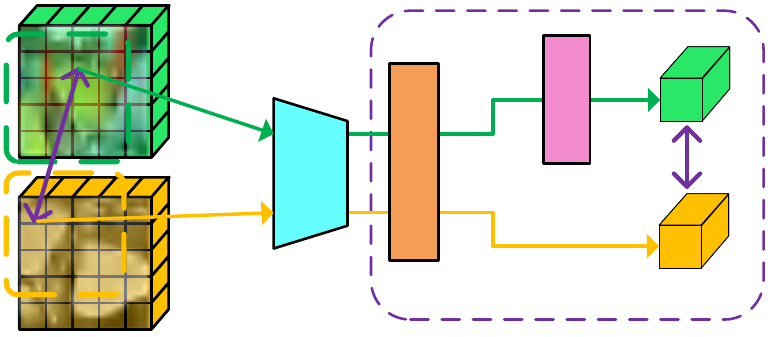}
		\put(38.8,27){\rotatebox{270}{Network}}
		\put(52,24){\rotatebox{270}{$\mathbf{M_p}$}}
		\put(72,33){\rotatebox{270}{$\mathbf{P}$}}
		\put(64,20){non-contrastive}
		\put(68,17){alignment}
		\put(45,27){$\tdZ_1$}
		\put(45,13){$\tdZ_2$}
		\put(58,27){$\tdV_1$}
		\put(58,13){$\tdV_2$}
		\put(87,39){$\mathbf{P}(\tdV_1)$}
		\put(88,5){$\tdV_2$}
		\put(90,22){$L_s$}
	\end{overpic} \vspace{-14pt}
	\caption{Illustration of non-contrastive pixel-to-pixel representation
		alignment.
		$\mathbf{M_p}$ is the projection layer that ensures less interference of
		pixel-level representation to the category representation.
		$\mathbf{P}$ is a pixel-level predictor for the asymmetric loss.
	}\label{fig:p2p}
\end{figure}
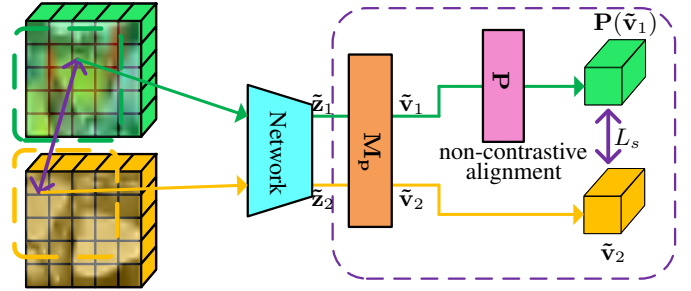

\begin{table}[t]
	\caption{Definition of frequently used symbols.}
	\vspace{-6pt}
	\centering
	\footnotesize
	\setlength{\tabcolsep}{0.3mm}
	\begin{tabular}{ccl}
		\toprule
		Symbols    & Dimensions/Type & Meaning                                         \\  \midrule
		$\featz$   & $L \sRes$       & output feature of one image.                    \\
		$\featz_k$ & $L \sRes$       & output feature of the $k$-th image.             \\
		$\featq_k$ & $(C+1)\sRes$    & pixel-level pseudo labels of the $k$-th image.  \\
		$\featy_k$ & $(C+1)\sRes$    & pixel-level GT labels of the $k$-th image.      \\
		$C$        & scalar          & number of major categories.                     \\
		$L$        & scalar          & number of dimensions of output feature.         \\
		$H$        & scalar          & height of output feature.                       \\
		$W$        & scalar          & width of output feature.                        \\
		$N$        & scalar          & number of images.                               \\
		$\Pooling$ & operation       & global average pooling over spatial dimensions. \\
		\bottomrule
	\end{tabular}
	\label{tab:symbols}
\end{table}

\myPara{Non-contrastive pixel-to-pixel representation alignment.}
Pixel-level shape-related representation aims to enhance
the feature discrimination ability in pixel-level,
\ie pixels within the same category or from the same image position
of different views should have consistent representations, and vice versa.
We observe that most existing pixel-level methods perform worse
than image-level methods on the LUSS task.
We argue existing pixel-level methods focus too much on the pixel-level
distinction,
thus resulting in semantic variation among pixels within the same instance.
To avoid the side-effect of pixel-level representation to instance-level
category representation,
we propose a non-contrastive pixel-to-pixel representation alignment strategy
that aligns the features from the same image position of different views
but avoids pushing features from other positions away.

As shown in~\figref{fig:p2p},
given the feature pair predicted from two views of the image,
we extract features $(\tdZ_1, \tdZ_2)$ of the overlapped pixels
and get the pixel-level embedding pairs ($\tdV_1$,$\tdV_2$)
by the projection $\tdV = \mathbf{M_p(\tdZ)}$,
where $\mathbf{M_p}$ is the pixel-level multi-layer projection (MLP) layer composed of the two
conv1$\times$1 layers and activation layers.
We show in~\secref{sec:ablation} that the projection
$\mProj$ ensures less interference
of pixel-level representation to the category representation.
We utilize the pixel-to-pixel alignment to align the overlapped
pixel-level embeddings from two views using the asymmetric loss:
\begin{equation}\label{eq:pixel2pixel}
	L_{P2P} =
	L_s(\mathbf{P}(\tdV_1),\StopG(\tdV_2))+L_s(\StopG(\tdV_1),\mathbf{P}(\tdV_2)),
\end{equation}
where the projection $\mathbf{P}$ is a pixel-level MLP predictor,
$\StopG$ is the stop gradient operation to avoid the collapse of predictor
\cite{chen2020exploring},
and $L_s$ is a cosine similarity loss.
The proposed non-contrastive pixel-to-pixel alignment forms a
robust pixel-level representation across
different views and maintains the category representation ability.

\myPara{Deep-to-shallow supervision.}
The quality of low/mid-level representation,
\ie representation in early network layers, is proven critical
to vision tasks~\cite{Islam_2021_ICCV,kotar2021contrasting}.
Islam \etal~\cite{Islam_2021_ICCV} reveal representations
with rich low/mid-level semantics in early layers result in
quick adaptation to a new task.
Similarly, Kotar \etal~\cite{kotar2021contrasting} show the benefit of
high-quality low-level features learned with contrastive-based methods.
Most existing works optimize mid-level representation by the indirect
gradient back-propagation from the high-level of the network
\cite{He_2020_CVPR,chen2020simple,li2020prototypical,caron2020unsupervised}.
We observe that directly applying low/mid-level features for
representation learning
leads to sub-optimal performance as these features lack semantics.
Therefore, we propose a deep-to-shallow supervision strategy to
enhance the representation of low/mid-level features with
the guidance of high-quality high-level features.

\begin{figure}[t]
	\centering
	\small
	\begin{overpic}[width=\linewidth]{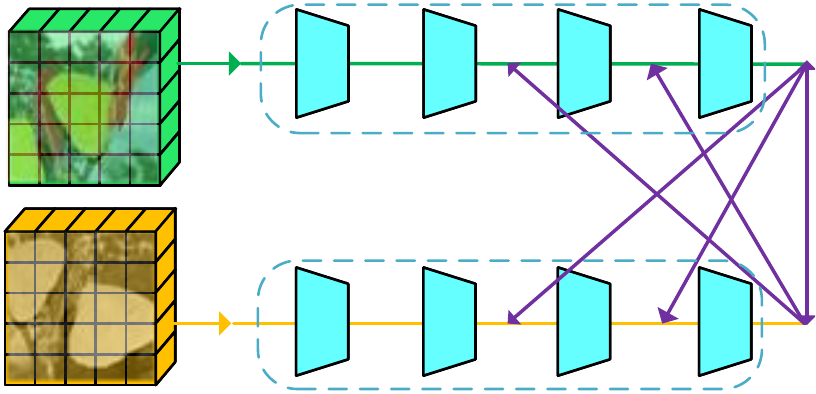}
		\put(95,4){$\mathbf{u}_2^{(4)}$}
		\put(77,4){$\mathbf{u}_2^{(3)}$}
		\put(60,4){$\mathbf{u}_2^{(2)}$}
		\put(45,4){$\mathbf{u}_2^{(1)}$}
		\put(95,42){$\mathbf{u}_1^{(4)}$}
		\put(77,42){$\mathbf{u}_1^{(3)}$}
		\put(60,42){$\mathbf{u}_1^{(2)}$}
		\put(45,42){$\mathbf{u}_1^{(1)}$}
	\end{overpic}
	\vspace{-14pt}
	\caption{Illustration of deep-to-shallow supervision.
		Purple lines denote the supervisions using loss $L_I$. $\mlp_I^{s}$ and $\mlp_K^{s}$
		are omitted for simplicity.
	}\label{fig:d2s}
\end{figure}
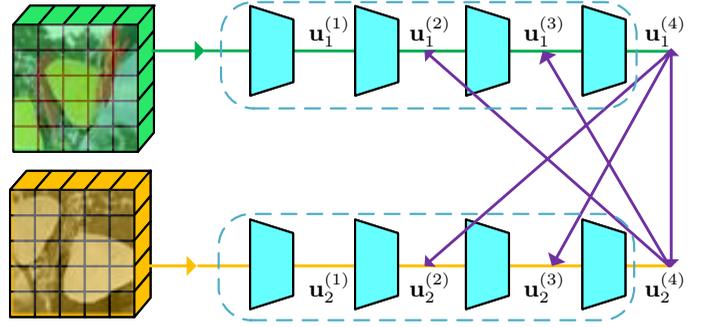

As shown in~\figref{fig:d2s},
given two views augmented from one image,
we obtain the feature pairs $(\featz_1^{(s)}, \featz_2^{(s)})$
from the $s$ stage of the network.
We mainly explore the effect of deep-to-shallow supervision on image-level for simplicity.
Given a network with four stages,
the image-level embeddings used for deep-to-shallow supervision are obtained as follows:
\begin{equation}
	\mathbf{u}_{i}^{(s)} =
	\begin{cases}
		\mlp^s_I (\Pooling(\featz_i^{(s)}))              & s = 4; \\
		\mlp^s_I (\Pooling(\mlp_K^{s}(\featz_i^{(s)}) )) & s < 4,
	\end{cases}
	\label{eq:embd_d2s}
\end{equation}
where $\Pooling$ is the global average pooling operation in the
spatial dimension,
$\mlp_I^{s}$ and $\mlp_K^{s}$ are the image-level/pixel-level
MLP layers of the stage $s$, respectively.
We observe that directly pooling the mid-level features causes the
representation collapse,
and adding $\mlp_K^{s}$ avoids this problem.
In the deep-to-shallow supervision strategy,
the embeddings from the last stage of one view are used to
supervise embeddings of another view from all stages:
\begin{equation}
	L_{D2S} =
	\frac{1}{|S|} \sum_j^{j \in S}
	{L_{I}(\mathbf{u}_1^{(4)}, \mathbf{u}_2^{(j)})} +
	\frac{1}{|S|} \sum_j^{j \in S}
	{L_{I} (\mathbf{u}_2^{(4)}, \mathbf{u}_1^{(j)})},
	\label{eq:d2sloss}
\end{equation}
where $S$ is a set containing stages used for deep-to-shallow supervision, and
$L_I$ is the image-level loss.
$L_I$ can be multiple definitions,
and we use the clustering loss~\cite{caron2020unsupervised}
as $L_I$ in our work.

\myPara{Training loss for representation learning.}
Our proposed pixel-to-pixel alignment and deep-to-shallow supervision can cooperate
with existing methods
to improve representation quality.
The summarized loss for the unsupervised representation learning step is written as:
\begin{equation}
	L_{sum} = L_{P2P} + L_{D2S} + L_{e},
	\label{eq:sumloss}
\end{equation}
where $L_{e}$ is the loss of existing methods,
\eg SwAV~\cite{caron2020unsupervised} and PixelPro~\cite{xie2020propagate}.

\subsection{Pixel-label Generation with Pixel-Attention}
\label{sec:pixel-att}

After representation learning, we obtain the features set
$\feaZ = \{\featz_k \in \Real^{L \sRes}, k \in [1,N] \}$
for all training images,
where $N$ is the number of images,
$L$, $H$, and $W$ are the number of dimensions, height, and width of
the output features.
We cluster $\feaZ$ to obtain $C$ generated categories
and assign generated categories to each pixel.
A straightforward way for label generation
is to cluster embeddings of all pixels in the training set,
which is too costly due to the large-scale data in LUSS,
\eg clustering training images of \ourdata~at pixel-level with
$7 \times 7$ resolution requires about \num{ 114 } hours.
An alternative is to use image-level features pooled on the
spatial dimension to save clustering costs.
However, many irrelevant embeddings are included in the pooled features,
hurting the clustering quality.

We observe that the learned features tend to focus on the regions
with more semantic meanings,
\ie pixels with more useful semantic information contribute more to the
convergence of unsupervised representation learning.
Based on this observation,
we propose a pixel-attention scheme to highlight meaningful semantic regions,
facilitating the pixel-level label generation with image-level features.
Specifically, we add a pixel-attention head at the output of models and
fine-tune it with representation learning losses to
filter out the less semantic meaningful regions.
Filtering features with pixel-attention reduces noise in
the pooled image-level embeddings, improving the clustering quality.
Also, the pixel-attention separates the semantic regions with less
meaningful regions,
generating more accurate object shapes during pixel-level label generation.
We give the implementation detail of pixel-attention in fine-tuning
and label generation steps.

\def\pAtt{\mathbf{c(z)}}

\myPara{Fine-tuning pixel-attention.}
Given the feature $\featz$ of one image predicted by the model,
representation learning methods \cite{caron2020unsupervised,He_2020_CVPR},
calculate losses with the pooled feature embeddings
$\mlp_I (\Pooling(\featz))$,
where $\mlp_I$ is the image-level MLP layer.
The pooling operation treats all pixels equally,
inevitably introducing noises to image-level embeddings
as not all pixels represent meaningful semantics.
Our pixel-attention is defined as:
\begin{equation}\label{eq:pixel_att}
	\pAtt = \sigma (\mlp_A (\| \featz \|) + \theta),
\end{equation}
where $\mlp_A$ is the pixel-level MLP layer,
$\theta \in \Real^L$ are learnable parameters initialized with zero,
$\sigma$ is a sigmoid function to restrict output attention value,
and $\| \featz \|$ is the L2 normalization operation applied on
the channel dimension of feature $\featz$.
Each channel of $\featz$ has a corresponding pixel-attention map.
We multiply the pixel-attention to feature $\featz$ and
obtain the pixel-attention enhanced image-level embedding
$\mathbf{\hat{v}=M}_I (\Pooling(\mathbf{c(z) \cdot \| z} \|) )$.
During fine-tuning,
we detach gradients to the network and fine-tune the pixel-attention module
by optimizing representation loss calculated from $\mathbf{\hat{v}}$.
We observe that fine-tuning pixel-attention module with
clustering loss~\cite{caron2020unsupervised} results in
decent shape-related pixel-attention results (see~\figref{fig:att_vis}).

\renewcommand{\addImg}[1]{{\includegraphics[width=.0787\textwidth]{pixelattention/ILSVRC2012_val_000#1}}}
\begin{figure}[t]
	\centering
	\renewcommand{\arraystretch}{0.5}
	\setlength{\tabcolsep}{0.2mm}
	\begin{tabular}{ccccccc}
		\addImg{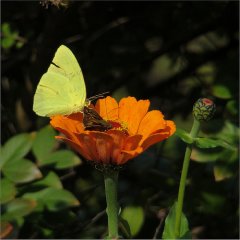}         &
		\addImg{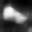} &
		\addImg{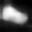} &
		\addImg{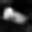} &
		\addImg{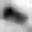} &
		\addImg{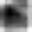} &
		\\
		\addImg{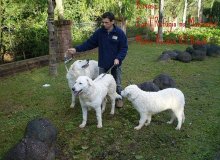}         &
		\addImg{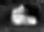} &
		\addImg{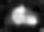} &
		\addImg{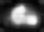} &
		\addImg{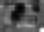} &
		\addImg{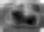} &
		\\
		\addImg{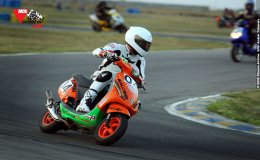}         &
		\addImg{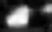} &
		\addImg{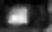} &
		\addImg{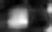} &
		\addImg{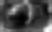} &
		\addImg{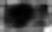} &
		\\
	\end{tabular}
	\caption{Visualization of pixel-attention maps of different channels.
		Most pixel-attention maps highlight the semantic regions,
		while a few channels highlight the background regions.
	}\label{fig:att_vis}
\end{figure}

\def\peilf{\mathbf{\hat{Z}}}

\myPara{Label generation with pixel-attention.}
Based on pixel-attention $\pAtt$,
we obtain the pixel-attention enhanced image-level feature
$\peilf = \{\peilf_k \in \Real^{L}, k \in [1,N] \}$,
where $\peilf_k = \Pooling(\pAtt_k \cdot \featz_k) $.
We conduct the k-means clustering over $\peilf$
to generate the cluster center $K \in \Real^{L \times C}$ of $C$ categories.
With the generated categories,
we need to assign pixel-level pseudo labels
$\feaQ=\{\featq_k \in \Real^{(C+1) \sRes}, k \in [1,N]\}$
to images.
We show in~\figref{fig:att_vis} the fine-tuned pixel-attention highlights
semantic regions of images.
Therefore, we extract regions with rich semantic information
based on pixel-attention:
\begin{equation}
	\mathbf{d(z)} = \begin{cases}
		0 & \frac{1}{L}\sum_i^{0 \le i < L} \pAtt_i  < \tau;   \\
		1 & \frac{1}{L}\sum_i^{0 \le i < L} \pAtt_i  \ge \tau,
	\end{cases}
	\label{eq:pixel_att_region}
\end{equation}
where $\tau$ is a pre-defined threshold between major categories and
the `other' category,
and regions with pixel-attention below $\tau$ are traded as the `other' category.
For each pixel in the region of major categories,
we assign one category in the cluster center $K$ that has
the minimal distance to the feature embedding of the pixel.

\subsection{Fine-tuning and Inference}

During the fine-tuning step,
we load the weights pre-trained with representation learning
and add a conv 1$\times$1 layer with $L\times (C+1)$ channels as the segmentation head.
The output features $\feaY = \{\featy_k \in \Real^{(C+1)\sRes}, k \in [1,N]\}$
from this head are supervised with $\feaQ$ to fine-tune the model
using cross-entropy loss.
During inference, the LUSS model acts like a fully supervised semantic
segmentation model.
For each pixel embedding $\mathbf{w} \in \Real^{C+1}$ in $\featy_k$,
we get the segmentation labels as follow:
\begin{equation}
	\mathbf{w} = {\arg\max}_{i\in [1,C+1]}(\mathbf{w}_{i}).
\end{equation}

\section{Experiments and Analysis}

\subsection{Implementation Details}
\label{sec:training_detail}

\myPara{Training details on the representation learning step.}
We use ResNet-18 network for the \ourdata$_{50}$ dataset
and utilize ResNet-50 for \ourdata$_{300}$ and \ourdata~datasets.
For a fair comparison, all networks are trained with a mini-batch size of \num{256}
for \num{ 200 } epochs in \ourdata$_{50}$
and \num{ 100 } epochs in \ourdata$_{300}$/\ourdata.

Our method cooperates with the image-level method
SwAV~\cite{caron2020unsupervised}
and the pixel-level method PixelPro~\cite{xie2020propagate}.
Following SwAV~\cite{caron2020unsupervised},
a LARS optimizer is used to update the network with a weight decay of
1e-6 and a momentum of 0.9.
The initial learning rate is 0.6 and gradually decays to 6e-6
with the cosine learning rate schedule.
For \ourdata$_{300}$ and \ourdata~datasets,
to make a fair comparison with other methods,
we only use two crops with the size of 224$\times$224 for training,
and the multi-crop training strategy~\cite{caron2020unsupervised}
is not applied.
Same as SwAV, a queue with the length of \num{ 3840 } is used beginning from \num{ 15 } epochs,
and the prototypes for clustering are frozen before \num{ 5005 } iterations.
When training on \ourdata${_{50}}$, the queue is set to 2048,
and prototypes are frozen before \num{ 1001 } iterations to ease the convergence.
We use the multi-crop training strategy containing 6 crops with
the size of 96$\times$96
and two crops with the size of 224$\times$224 for training
on \ourdata${_{50}}$.
When cooperating with PixelPro~\cite{xie2020propagate},
the training schemes are consistent with the official setting.
We train the network with the initial learning rate of 1.0 using a
LARS optimizer.
After five warm-up epochs,
the learning rate gradually decays to 1e-6 with the cosine learning
rate schedule.

\myPara{Training details on the fine-tuning step.}
To generate pixel-level labels,
we first fine-tune the pixel-attention module for \num{ 20 } epochs
while fixing the model parameters trained in the unsupervised
representation learning step.
We apply the clustering loss~\cite{caron2020unsupervised} to fine-tune
the pixel-attention module by default.
The training strategy remains the same as that in the
representation learning step.

The representation learning losses are removed in the fine-tuning step,
and an extra cross-entropy loss is added to supervise the segmentation head.
We load the model weights pre-trained on the representation learning step
and fine-tune them for \num{ 20 } epochs.
We train the network using a LARS optimizer with a weight decay of 1e-6,
a mini-batch size of 256, and a momentum of 0.9.
The initial learning rate is 0.6 and gradually decays
to 6e-6 with the cosine learning rate schedule.

\newcommand{\tRows}[1]{\multirow{2}*{#1}}
\newcommand{\tCols}[1]{\multicolumn{2}{c}{#1}}
\def\MDC{MDC\cite{caron2018deep,cho2021picie}}
\def\PiCIE{PiCIE\cite{cho2021picie}}
\def\MaskCon{MaskCon\cite{van2020unsuperv}}

\begin{table*}[t]
	\caption{Comparison of our proposed LUSS method and existing
		USS methods on the \ourdata~dataset
		using the fully unsupervised evaluation protocol.
		Test mIoU under different object sizes are provided.
		\dag~means train model for \num{ 200 } epochs from scratch.
		PASS$_{s/p}$ denotes using
		SwAV~\cite{caron2020unsupervised} and PixelPro~\cite{xie2020propagate}
		as the $L_e$ in~\eqref{eq:sumloss}, respectively.
		\textit{S} means the method use saliency maps.
		\textit{I} means initialization with supervised ImageNet$_{1k}$ pre-training.
		By default, the `other' category is used to calculate
		mIoU and b-mIoU. We also give the performance
		without using the `other' category in the supplementary.
	}
	\vspace{10pt}
	\centering
	\footnotesize
	\setlength{\tabcolsep}{2.0mm}
	\begin{tabular}{lccccccccccccccccc}
		\toprule
		\tRows{LUSS}                 & \tRows{Pior}        & \tCols{mIoU}             & \tCols{b-mIoU}             &
		\tCols{Img-Acc}              & \tCols{F$_{\beta}$} & \multicolumn{4}{c}{mIoU} & \multicolumn{4}{c}{b-mIoU}                                                                                                                                                                                       \\
		                             &                     & val                      & test                       & val        & test       & val        & test       & val        & test       & S.         & M.S.       & M.L.       & L.         & S.         & M.S.       & M.L.       & L.         \\ \midrule
		\multicolumn{17}{c}{\ourdata$_{50}$}                                                                                                                                                                                                                                                             \\ \midrule
		\MDC                         & -
		                             & 4.0                 & 3.6                      & 1.4                        & 1.2        & 14.9       & 13.4       & 31.6       & 31.3       & 0.4        & 2.6        & 3.8        & 4.9        & 0.2        & 1.1        & 1.4        & 1.5                     \\
		\MDC                         & \textit{I}
		                             & 14.6                & 14.3                     & 3.1                        & 3.1        & 44.8       & 40.8       & 33.2       & 32.6       & 2.6        & 10.9       & 14.6       & 19.1       & 0.9        & 2.2        & 3.2        & 4.7                     \\
		\PiCIE                       & -
		                             & 5.0                 & 4.5                      & 1.8                        & 1.6        & 15.8       & 14.0       & 14.6       & 32.2       & 0.2        & 3.1        & 5.0        & 5.3        & 0.2        & 1.2        & 1.7        & 1.9                     \\
		\PiCIE                       & \textit{I}
		                             & 17.8                & 17.6                     & 3.7                        & 4.0        & 45.0       & 44.0       & 32.1       & 31.6       & 4.4        & 13.1       & 20.1       & 23.1       & 1.0        & 2.7        & 4.4        & 5.8                     \\
		\MaskCon                     & \textit{S}
		                             & 24.6                & 24.2                     & 15.6                       & 15.1       & 47.9       & 47.6       & 65.7       & 66.2       & 12.2       & 25.6       & 24.7       & 20.4       & 10.1       & 17.0       & 14.5       & 10.6                    \\
		\MaskCon\dag                 & \textit{S}
		                             & 13.9                & 10.5                     & 8.5                        & 10.5       & 30.2       & 22.4       & 62.6       & 62.3       & 2.5        & 2.1        & 1.7        & 1.7        & 2.4        & 6.3        & 6.5        & 5.7                     \\
		\OursS                       & -
		                             & 29.2                & 29.3                     & 7.6                        & 7.4        & {\bf 66.2} & {\bf 65.5} & 49.0       & 49.0       & 6.6        & 25.0       & 33.2       & 32.6       & 3.3        & 6.2        & 8.1        & 9.5                     \\
		\OursP                       & -
		                             & 32.4                & 32.0                     & 7.2                        & 7.2        & 62.9       & 64.1       & 48.7       & 47.9       & 9.7        & 26.2       & 36.5       & 40.5       & 5.1        & 5.8        & 7.8        & 10.4                    \\
		\OursP + RC~\cite{ChengPAMI} & \textit{S}          & 42.6                     & 42.1                       & 17.5       & 17.7       & 58.8       & 61.8       & 62.1       & 61.3       & 17.0       & 38.6       & {\bf 45.5} & {\bf 43.7} & 11.2       & 17.2       & 19.0       & {\bf 17.1} \\
		\OursP + Sal                 & \textit{S}          & {\bf 43.3}               & {\bf 42.3}                 & {\bf 20.4} & {\bf 20.2} & 64.6       & 65.2       & {\bf 70.0} & {\bf 69.9} & {\bf 19.0} & {\bf 41.7} & 45.1       & 38.3       & {\bf 14.7} & {\bf 22.6} & {\bf 20.6} & 15.3       \\
		\midrule
		\multicolumn{17}{c}{\ourdata$_{300}$}                                                                                                                                                                                                                                                            \\
		\midrule
		\OursP                       & -
		                             & 16.6                & 16.0                     & 4.4                        & 4.2        & 34.7       & 32.8       & 34.4       & 34.3       & 2.8        & 12.0       & 16.4       & 21.7       & 1.4        & 3.2        & 3.9        & 6.4                     \\
		\OursS                       & -
		                             & {\bf 18.0}          & {\bf 18.1}               & {\bf 5.2}                  & {\bf 5.2}  & {\bf 43.9} & {\bf 42.6} & {\bf 47.6} & {\bf 47.5} & {\bf 4.2}  & {\bf 13.6} & {\bf 19.5} & {\bf 23.5} & {\bf 2.1}  & {\bf 4.2}  & {\bf 5.5}  & {\bf 7.1}               \\
		\midrule
		\multicolumn{17}{c}{\ourdata}                                                                                                                                                                                                                                                                    \\
		\midrule
		\OursP                       & -
		                             & 7.3                 & 6.6                      & 2.4                        & 2.1        & 19.9       & 18.0       & 34.8       & 34.6       & 1.3        & 4.6        & 7.1        & 8.4        & 0.6        & 1.5        & 2.1        & 2.8                     \\
		\OursS                       & -
		                             & {\bf 11.5}          & {\bf 11.0}               & {\bf 3.8}                  & {\bf 3.5}  & {\bf 24.0} & {\bf 22.3} & {\bf 37.1} & {\bf 36.9} & {\bf 2.4}  & {\bf 8.3}  & {\bf 11.9} & {\bf 13.4} & {\bf 1.3}  & {\bf 3.0}  & {\bf 3.8}  & {\bf 4.3}               \\
		\bottomrule
	\end{tabular}
	\label{tab:unsup_com}
\end{table*}

\subsection{Comparison with USS Methods}

In this section,
we evaluate the performance of the proposed LUSS method
on the \ourdata~dataset using the fully unsupervised evaluation protocol.
\tabref{tab:unsup_com} shows our method achieves reasonable performance on
the large-scale data.
The visualization shown in~\figref{fig:vis} indicates that
unsupervised semantic segmentation with the large-scale dataset is achievable.

\myPara{Comparison with unsupervised semantic segmentation methods.}
Existing unsupervised semantic segmentation (USS) methods are
designed for relatively small scale data,
thus cannot be directly used on the full-scale \ourdata~dataset
due to the training time limit.
Therefore, we compare our LUSS method with existing USS methods
on the \ourdata$_{50}$ subset,
as shown in~\tabref{tab:unsup_com}.
All methods trained on \ourdata$_{50}$
utilize the ResNet-18 network for a fair comparison.
The comparison is not strictly fair because some existing USS methods
are not trained in fully unsupervised settings.
For example, MDC~\cite{caron2018deep} and PiCIE~\cite{cho2021picie}
initialize models with supervised ImageNet$_{1k}$ pre-trained weights.
These two methods suffer from large performance drops
when using MoCo~\cite{He_2020_CVPR} pre-trained weights,
indicating that supervised pre-training is a vital step.
MaskContrast~\cite{van2020unsuperv}
is initialized with the MoCo pre-trained weights
and trained with extra saliency maps as supervision.
There is a large performance loss when this model is trained from scratch.
In contrast, our LUSS method is trained from scratch
with no direct/indirect human supervision.
Our method includes the proposed representation learning strategy,
label generation approach, and fine-tuning scheme.
To validate the generalizability of our method,
we implement our method based on two representation learning methods,
\ie \SwAV~and \PixelPro.
Our method outperforms existing USS methods with a clear margin in mIoU.
Benefiting from extra saliency maps,
MaskContrast has a higher F$_\beta$ than our method.
Using the same saliency maps, our saliency-enhanced method
clearly outperforms MaskContrast in F$_\beta$
and achieves much higher mIoU.
Note that saliency maps in~\cite{van2020unsuperv} are not strictly unsupervised
version because the supervised ImageNet pretraining weights are used.
When using saliency maps from the fully unsupervised method RC~\cite{ChengPAMI},
our method still achieves competitive performance.
We also implement other USS methods, \eg IIC~\cite{ji2019invariant}.
However, as these methods are designed for semantic segmentation
under several categories,
they fail to converge on the \ourdata$_{50}$ dataset.

\newcommand{\addFig}[1]{{\includegraphics[height=.065\textwidth,width=.075\textwidth]{figures/results/ILSVRC2012_val_000#1.jpg}}}
\begin{figure}[t]
	\centering
	\renewcommand{\arraystretch}{0.2}
	\setlength{\tabcolsep}{0.2mm}
	\begin{tabular}{cccccccc}
		\addFig{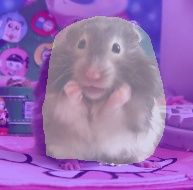} &
		\addFig{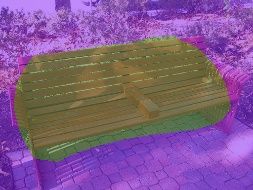} &
		\addFig{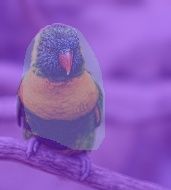} &
		\addFig{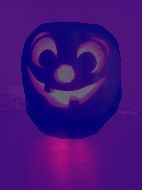} &
		\addFig{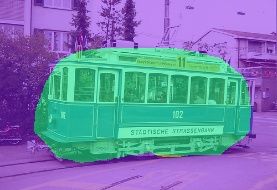} &
		\addFig{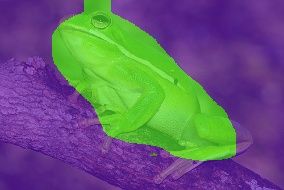}   \\
		\addFig{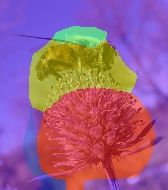} &
		\addFig{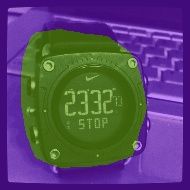} &
		\addFig{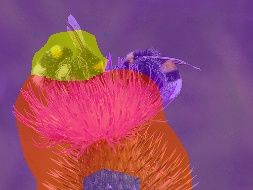} &
		\addFig{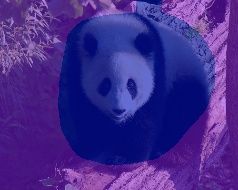} &
		\addFig{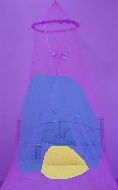} &
		\addFig{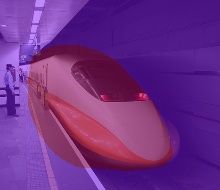}   \\
		\addFig{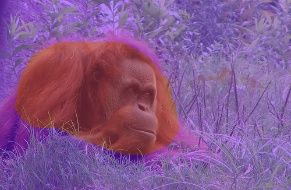} &
		\addFig{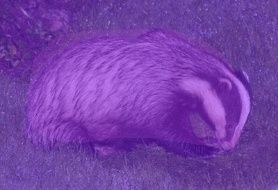} &
		\addFig{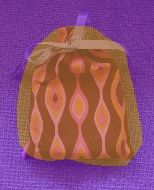} &
		\addFig{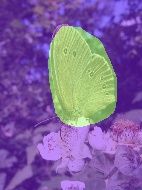} &
		\addFig{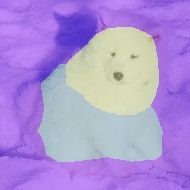} &
		\addFig{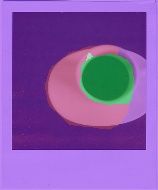}   \\
		\addFig{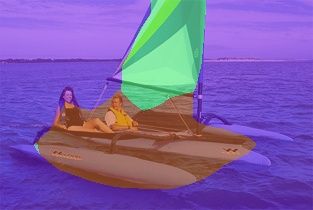} &
		\addFig{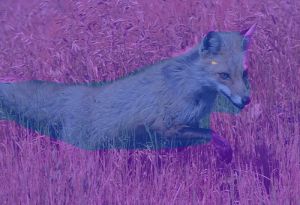} &
		\addFig{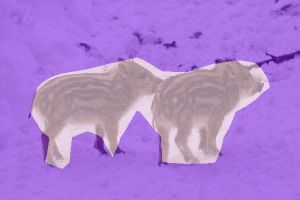} &
		\addFig{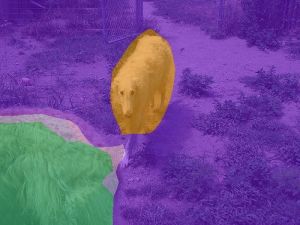} &
		\addFig{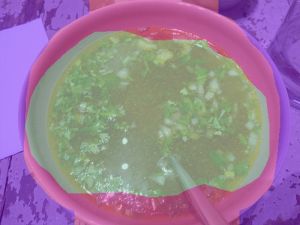} &
		\addFig{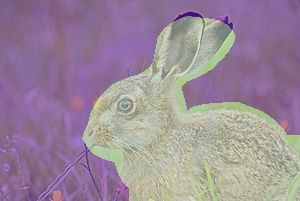}   \\
		\addFig{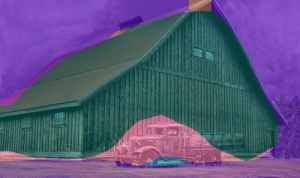} &
		\addFig{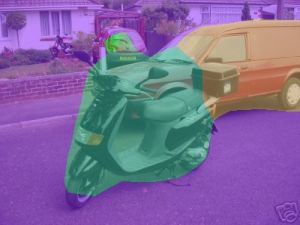} &
		\addFig{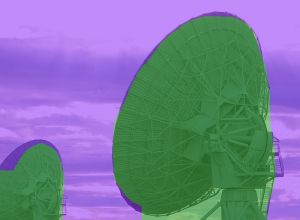} &
		\addFig{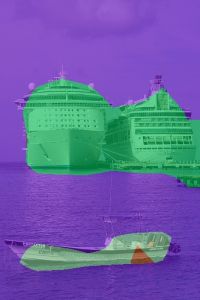} &
		\addFig{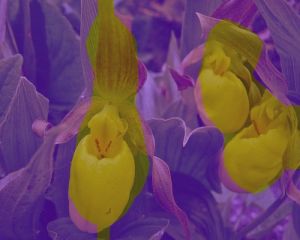} &
		\addFig{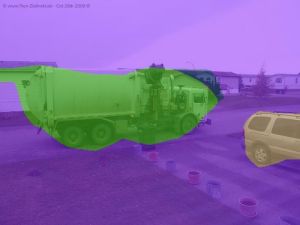}   \\
		\addFig{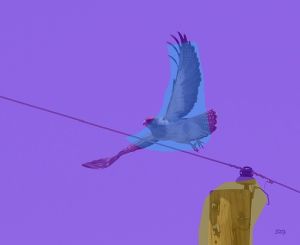} &
		\addFig{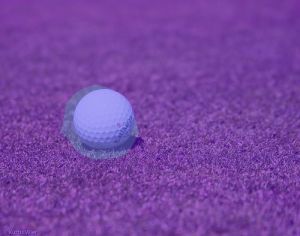} &
		\addFig{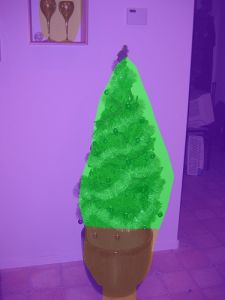} &
		\addFig{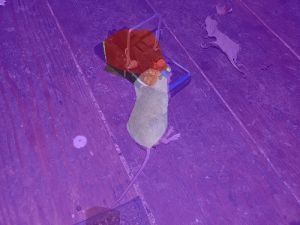} &
		\addFig{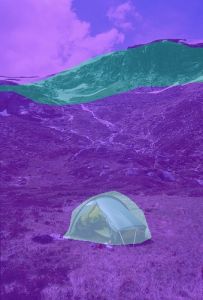} &
		\addFig{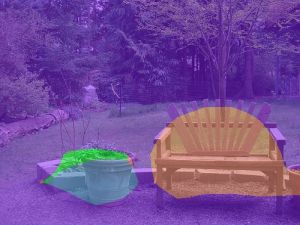} &
		\\
	\end{tabular}
	\caption{Visualization of unsupervised semantic segmentation results. Last three rows are trained with saliency prior information during label generation, showing better shape quality. }
	\label{fig:vis}
\end{figure}

\myPara{Performance of different object sizes.}
As introduced in~\secref{sec:dataset_state},
the \ourdata~dataset is divided into multiple groups by object size.
We evaluate the test mIoU under different object sizes,
as shown in~\tabref{tab:unsup_com}.
The performance of small objects is worse than large objects
in mask and boundary mIoU,
indicating that small objects need a model with a
more precise pixel-level representation and segmentation ability.
Note that performance in boundary mIoU has smaller gaps of different object sizes than mask mIoU since the boundary mIoU is more robust
to object size changes.

\myPara{Difference between different data scales.}
As shown in~\tabref{tab:unsup_com},
we train our method on \ourdata$_{50}$,
\ourdata$_{300}$, and \ourdata~datasets.
The performance scores drop with the growth of the data scale,
showing the great challenge of the large-scale data to
unsupervised semantic segmentation.
We observe that PixelPro based method outperforms the SwAV based method
on the \ourdata$_{50}$ dataset,
but the SwAV based method achieves better performance on
\ourdata$_{300}$ and \ourdata~datasets.
We conclude that different data scales prefer different representation
learning strategies.

To evaluate the performance gap between large and small datasets,
we train models on large sets and evaluate models
on small sets, as shown in~\tabref{tab:unsup_com_largetrain_smalltest}.
Models trained on large sets are inferior to models trained on small sets.
Testing on the ImageNet-S$_{50}$ set,
the model trained on ImageNet-S$_{50}$ achieves the best performance for all metrics,
while the model trained with ImageNet-S$_{919}$ has the worst scores.
A similar trend is also observed when evaluating on ImageNet-S$_{300}$ set.
These results indicate that training unsupervised models on larger datasets
is harder than on small datasets, showing the great challenge of large-scale data.
However, the performance gaps are relatively small, which may be closed by stronger future methods.

\begin{table}[t]
	\caption{Training models on the large set and evaluating models on the small subsets of \ourdata~with fully unsupervised protocol using \OursS~ method.
	}
	\vspace{10pt}
	\centering
	\footnotesize
	\setlength{\tabcolsep}{2.5mm}
	\begin{tabular}{lcccccccccc}
		\toprule
		\tRows{Training set} & \tCols{mIoU} & \tCols{Img-Acc} & \tCols{F$_{\beta}$}                                        \\
		                     & val          & test            & val                 & test       & val        & test       \\ \midrule
		\multicolumn{7}{c}{Testing on \ourdata$_{50}$}                                                                     \\ \midrule
		\ourdata$_{50}$
		                     & {\bf 29.2}   & {\bf 29.3}      & {\bf 66.2}          & {\bf 65.5} & {\bf 49.0} & {\bf 49.0} \\
		\ourdata$_{300}$     & 27.8         & 27.4            & 65.2                & 63.3       & 38.6       & 36.0       \\
		\ourdata             & 24.1         & 23.0            & 61.3                & 57.8       & 31.3       & 28.9       \\
		\midrule
		\multicolumn{7}{c}{Testing on \ourdata$_{300}$}                                                                    \\
		\midrule
		\ourdata$_{300}$
		                     & {\bf 18.0}   & {\bf 18.1}      & {\bf 43.9}          & {\bf 42.6} & {\bf 47.6} & {\bf 47.5} \\
		\ourdata             & 16.4         & 16.6            & 39.3                & 37.2       & 36.8       & 36.0       \\
		\bottomrule
	\end{tabular}
	\label{tab:unsup_com_largetrain_smalltest}
\end{table}

\begin{table}
	\centering
	\caption{Ablation of the proposed P2P alignment and D2S supervision
		representation learning strategy.
		All models are trained with \num{ 100 } epochs.
		D2S3 and D2S32 mean supervising stage \num{ 3 } and stage 3-2 of the network, respectively.
	}
	\subtable[Ablation on LUSS using distance matching evaluation protocol.]{
		\setlength{\tabcolsep}{2.67mm}
		\begin{tabular}{llccccccc}
			\toprule
			\tRows{\ourdata$_{300}$} & \tCols{mIoU} & \tCols{Img-Acc} & \tCols{$F_{\beta}$}                      \\
			                         & val          & test            & val                 & test & val  & test \\ \midrule
			\SwAV                    & 22.4         & 22.6            & 57.4                & 57.5 & 63.5 & 63.7 \\
			+P2P                     & 24.8         & 24.8            & 58.4                & 58.5 & 64.5 & 64.8 \\
			+P2P-D2S3                & 25.1         & 25.2            & 57.3                & 57.5 & 65.0 & 65.2 \\
			+P2P-D2S32               & 24.8         & 24.9            & 56.8                & 56.6 & 65.7 & 66.0 \\ \midrule
			\PixelPro                & 15.5         & 15.8            & 44.0                & 44.3 & 62.4 & 62.6 \\
			+Clustering Loss         & 20.8         & 21.3            & 52.0                & 52.1 & 61.5 & 62.1 \\
			+P2P                     & 21.3         & 22.0            & 52.2                & 52.8 & 61.5 & 62.1 \\
			+P2P-D2S3                & 22.2         & 22.8            & 53.2                & 53.1 & 62.2 & 62.9 \\
			+P2P-D2S32               & 23.0         & 23.4            & 53.3                & 54.3 & 62.4 & 63.1 \\ \bottomrule
		\end{tabular}
		\label{tab:p2p-d2s-luss}
	}
	\subtable[Ablation of transfer learning on downstream tasks.]{
		\setlength{\tabcolsep}{1.05mm}
		\begin{tabular}{llcccccccc}	\toprule
			\tRows{\ourdata$_{300}$} & \thCols{COCO SEG}
			                         & \thCols{COCO DET} & VOC SEG &                                  \\ \cline{2-8}
			                         & AP                & AP50    & AP75 & AP   & AP50 & AP75 & mIoU \\ \midrule
			\SwAV                    & 32.4              & 52.1    & 34.6 & 35.5 & 54.9 & 38.6 & 68.9 \\
			+P2P                     & 32.8              & 52.5    & 34.9 & 36.0 & 55.4 & 39.1 & 70.4 \\
			+P2P-D2S3                & 33.5              & 53.4    & 35.8 & 36.7 & 56.4 & 39.4 & 70.8 \\
			+P2P-D2S32               & 33.8              & 53.7    & 36.2 & 37.2 & 56.6 & 40.6 & 70.8 \\ \midrule
			\PixelPro                & 34.7              & 54.8    & 37.2 & 38.2 & 57.5 & 41.7 & 72.8 \\
			+Clustering Loss         & 34.9              & 55.2    & 37.3 & 38.4 & 58.1 & 41.9 & 73.3 \\
			+P2P                     & 35.3              & 55.9    & 37.9 & 38.9 & 58.6 & 42.4 & 72.3 \\
			+P2P-D2S3                & 35.3              & 55.9    & 37.6 & 38.8 & 58.6 & 42.3 & 73.9 \\
			+P2P-D2S32               & 35.7              & 56.6    & 38.3 & 39.4 & 59.1 & 43.1 & 75.1 \\
			\bottomrule
		\end{tabular}
		\label{tab:p2p-d2s-downstream}
	}
\end{table}

\subsection{Ablation}
\subsubsection{Representation Learning}
\label{sec:ablation}

In this section,
we benchmark our proposed and some existing unsupervised representation
learning methods on the LUSS task.
We conduct experiments on the \ourdata$_{300}$ dataset to save
computational cost unless otherwise stated.
To avoid the influence of fine-tuning step in LUSS,
we apply the distance matching protocol for LUSS evaluation
as introduced in~\secref{sec:protocols}.

\myPara{Ablation of the proposed representation learning strategy.}
We implement the proposed non-negative pixel-to-pixel (P2P) alignment
and deep-to-shallow (D2S) supervision on top of \SwAV ~and \PixelPro.
\tabref{tab:p2p-d2s-luss} shows that PixelPro performs much worse than
SwAV due to the missing category-related representation ability
needed by LUSS.
Therefore, we add the clustering loss~\cite{caron2020unsupervised}
to PixelPro to form a reasonable baseline model for LUSS.
As shown in~\tabref{tab:p2p-d2s-luss},
our method improves the SwAV and PixelPro with 2.6\% and 7.6\% in
test mIoU on the \ourdata$_{300}$ set, respectively.
Specifically, the P2P alignment
has a gain of 2.2\% in test mIoU compared to the image-level method SwAV,
and it also improves the clustering-loss enhanced PixelPro by 0.5\%
in test mIoU.
The D2S supervision brings the further gain of 0.4\% and 1.4\%
over SwAV and PixelPro based baselines, respectively.
In summary, the P2P alignment effectively enhances the pixel-level
representation of image-level methods,
and D2S supervision enriches the instance-level category representation
of pixel-level methods.
The P2P alignment and D2S supervision still improve the methods designed
for pixel-level and image-level representation, respectively,
showing the robustness of the proposed strategies.
As shown in~\tabref{tab:benchmarks},
our proposed representation learning strategy also outperforms baselines
on the \ourdata~dataset.

\myPara{Non-negative pixel-to-pixel alignment.}
We utilize the non-negative P2P alignment to enhance
the pixel-level representation without hurting the instance-level category representation.
We also compare different pixel-level alignment strategies, including
clustering, contrastive, and non-contrastive types.
We set pixels at the same position of two views as positive pairs and
other pixels as negative pixels.
As shown in~\tabref{tab:p2p_mode},
both three pixel-level alignment strategies have higher $F_{\beta}$ compared to the baseline,
showing improved shape representation quality.
However, due to the semantic variation among pixels in the same object,
clustering and contrastive losses suffer from the performance drop on mIoU and Img-Acc.
In contrast, the proposed non-negative P2P alignment has performance gains over
the baseline in mIoU and Img-Acc, due to maintaining the representation constancy of pixels belonging to the same semantic instance.
We also analyze the effectiveness of the projection $\mProj$ in~\tabref{tab:proj_in_p2p}.
The P2P alignment with projection $\mProj$ achieves better Img-Acc
because $\mProj$ ensures less interference
of pixel-level representation to the category-related representation.

\begin{table}
	\caption{Ablation about the P2P alignment and D2S supervision strategy on the \ourdata$_{300}$ testing set using
		distance matching evaluation protocols.}
	\centering
	\subtable[Different loss types for P2P alignment.]{
		\small
		\centering
		\setlength{\tabcolsep}{4.0mm}
		\begin{tabular}{lccccc}
			\toprule
			\ourdata$_{300}$     & mIoU & Img-Acc & $F_{\beta}$ \\ \midrule
			SwAV baseline        & 22.6 & 57.5    & 63.7        \\
			+Clustering P2P      & 21.2 & 51.8    & 66.4        \\
			+Contrastive P2P     & 18.0 & 46.4    & 64.6        \\ \midrule
			+Non-contrastive P2P & 24.8 & 58.5    & 64.8        \\
			\bottomrule
		\end{tabular}
		\label{tab:p2p_mode}
	}
	\subtable[Effect of projections $\mlp$ in P2P alignment.]{
		\small
		\setlength{\tabcolsep}{4.3mm}
		\begin{tabular}{lccccc}
			\toprule
			\ourdata$_{300}$     & mIoU & Img-Acc & $F_{\beta}$ \\ \midrule
			SwAV baseline        & 22.6 & 57.5    & 63.7        \\ \midrule
			P2P without $\mProj$ & 24.6 & 57.1    & 64.9        \\
			P2P with $\mProj$    & 24.8 & 58.5    & 64.8        \\
			\bottomrule
		\end{tabular}
		\label{tab:proj_in_p2p}
	}
	\subtable[Deep-to-shallow versus same-stage supervisions in D2S supervision.]{
		\small
		\setlength{\tabcolsep}{3.8mm}
		\begin{tabular}{lccccc}
			\toprule
			\ourdata$_{300}$        & mIoU & Img-Acc & $F_{\beta}$ \\ \midrule
			PixelPro+P2P (baseline) & 22.0 & 52.8    & 62.1        \\ \midrule
			+same-stage sup.        & 22.6 & 52.9    & 63.1        \\
			+deep-to-shallow sup.   & 23.4 & 54.3    & 63.1        \\
			\bottomrule
		\end{tabular}
		\label{tab:d2s_type}
	}
	\subtable[Same-view versus cross-view supervisions in D2S supervision.]{
		\small
		\setlength{\tabcolsep}{3.8mm}
		\begin{tabular}{lccccc}
			\toprule
			\ourdata$_{300}$        & mIoU & Img-Acc & $F_{\beta}$ \\ \midrule
			PixelPro+P2P (baseline) & 22.0 & 52.8    & 62.1        \\ \midrule
			+same-view sup.         & 23.1 & 53.9    & 63.2        \\
			+cross-view sup.        & 23.4 & 54.3    & 63.1        \\
			\bottomrule
		\end{tabular}
		\label{tab:d2s_crossview}
	}
\end{table}

\begin{table}[t]
	\centering
	\small
	\caption{
		Performance comparison of unsupervised representation learning methods
		and our proposed representation learning enhancement methods using
		distance matching evaluation protocol.
		PASS$_{s/p}$ denotes using
		SwAV~\cite{caron2020unsupervised} and PixelPro~\cite{xie2020propagate}
		as the $L_e$ in~\eqref{eq:sumloss}, respectively.
		All models are trained with \num{ 100 } epochs.
		Supervised means initializing the model with image-level supervised pre-training.
	}\setlength{\tabcolsep}{2.1mm}
	\begin{tabular}{lcccccccc} \toprule
		\tRows{LUSS} & \tCols{mIoU} & \tCols{Img-Acc} & \tCols{$F_{\beta}$}                      \\
		             & val          & test            & val                 & test & val  & test \\ \midrule
		\multicolumn{7}{c}{\ourdata$_{300}$}                                                     \\ \midrule
		Supervised   & 33.8         & 33.9            & 80.4                & 81.5 & 60.0 & 60.0 \\ \midrule
		Contrastive                                                                              \\  \midrule
		\SimCLR      & 12.5         & 12.6            & 37.7                & 38.4 & 63.7 & 64.0 \\
		\MoCovT      & 12.4         & 12.4            & 40.3                & 40.3 & 64.1 & 64.4 \\
		\AdCo        & 21.1         & 21.5            & 55.1                & 54.8 & 64.9 & 65.5 \\ \midrule
		Non-contrastive                                                                          \\ \midrule
		\BYOL        & 13.4         & 13.4            & 38.3                & 38.0 & 64.0 & 64.4 \\
		\SimSiam     & 20.1         & 20.3            & 56.9                & 57.5 & 65.5 & 66.0 \\ \midrule
		Clustering                                                                               \\ \midrule
		\PCL         & 17.4         & 17.9            & 48.4                & 48.0 & 63.0 & 63.3 \\
		\SwAV        & 22.4         & 22.6            & 57.4                & 57.5 & 63.5 & 63.7 \\
		\OursS       & 25.1         & 25.2            & 57.3                & 57.5 & 65.0 & 65.2 \\	\midrule
		Pixel-level                                                                              \\ \midrule
		\DenseCL     & 13.9         & 13.8            & 36.4                & 36.8 & 63.7 & 63.7 \\
		\PixelPro    & 15.5         & 15.8            & 44.0                & 44.3 & 62.4 & 62.6 \\
		\OursP       & 23.0         & 23.4            & 53.3                & 54.3 & 62.4 & 63.1 \\
		\midrule
		\multicolumn{7}{c}{\ourdata}                                                             \\	\midrule
		Supervised   & 30.0         & 29.8            & 75.9                & 76.6 & 58.7 & 58.7 \\
		\PixelPro    & 7.7          & 7.5             & 26.9                & 26.5 & 61.8 & 61.8 \\
		\OursP       & 9.8          & 9.8             & 29.4                & 29.6 & 61.1 & 61.3 \\
		\SwAV        & 15.1         & 15.1            & 43.5                & 43.3 & 64.2 & 64.3 \\
		\OursS       & 15.6         & 15.6            & 43.1                & 42.9 & 64.3 & 64.6 \\
		\bottomrule
	\end{tabular}
	\label{tab:benchmarks}
\end{table}

\myPara{Deep-to-shallow supervision.}
The D2S supervision utilizes high-quality features from the last stage
to supervise early-stage features.
\tabref{tab:d2s_type} compares using features from the same or last stages
as supervision to shallow layers.
We observe that both settings have improvements over the baseline,
and the deep-to-shallow supervision outperforms the same-stage supervision
in mIoU and Img-Acc.
By default, we use the deep features from one view to supervising shallow features
from another view.
In~\tabref{tab:d2s_crossview},
we study the effects of applying D2S supervision on features belonging to the same view.
Cross-view supervision is slightly better than same-view supervision.
We observe that the training loss of same-view supervision is lower than cross-view supervision.
We conclude that the same-view supervision over-fits to a sub-optimal solution, hurting the evaluation performance.
The D2S supervises multiple features from different early stages of the network.
As shown in~\tabref{tab:p2p-d2s-luss}, we study the effects of supervising
different stages on SwAV and PixelPro based methods.
We observe that different methods require different stages to
get optimal results, \eg supervising stage 3-2 is worse than supervising stage \num{ 3 } in SwAV,
but PixelPro benefits from more supervision to the early stages.
We choose the stages for D2S supervision by ablation study.

\myPara{Benchmarking unsupervised learning methods.}
To analyze the representation ability of unsupervised learning methods on the LUSS task,
we categorize and benchmark some representative methods,
including contrastive, non-contrastive, clustering, and pixel-level methods.
As shown in~\tabref{tab:benchmarks},
image-level methods have a clear advantage over pixel-level methods on mIoU, image-level accuracy, and $F_{\beta}$.
Pixel-level methods focus too much on the pixel-level distinction,
resulting in semantic variation among pixels within the same instance, \ie
one instance contains multiple categories.
In comparison, image-level methods provide constant instance-level category-related representation
as these losses encourage distinguishing among images.
However, pixel-level representation is vital to the LUSS task
as our proposed non-contrastive P2P alignment method has a
considerable gain over the image-level method SwAV.
We observe that the clustering methods outperform the contrastive and non-contrastive methods
in image-level accuracy but have worse performance on shape-related $F_{\beta}$.
The clustering strategy encourages stronger category-related representations with category centroids
than contrastive and non-contrastive methods.
But due to all pixels of one image in clustering methods being close to category centroids,
the representation difference between major and other categories is weakened.
The image-level supervised method has better category centroids than the clustering method,
and it also has worse $F_{\beta}$ than clustering methods.
These results explain why clustering methods have worse $F_{\beta}$.

\myPara{What role does the category play in the LUSS task?}
To answer this question, we use the models trained with image-level supervision as the baseline.
As shown in~\tabref{tab:benchmarks},
the supervised model performs better than the unsupervised models in mIoU.
In addition, it outperforms unsupervised models in terms of image-level accuracy by a large margin.
In contrast, the performance in shape-related metric, \ie $F_\beta$, is worse than most unsupervised methods.
These results show that category features indeed facilitate the LUSS task.
However, shape features cannot be learned solely by category representation learning.

\def\Contrastive{Contrastive (MoCov2~\cite{He_2020_CVPR,chen2020mocov2})}
\def\NonContrastive{\Non-contrastive (SimSam~\cite{chen2020exploring})}
\def\Clustering{Clustering (SwAV~\cite{caron2020unsupervised})}
\newcommand{\tPow}[2]{$#1 \times 10^{#2}$}

\begin{table}
	\caption{Ablation about pixel-label generation and fine-tuning steps on the \ourdata$_{50}$ testing set
		using fully unsupervised evaluation protocol.}
	\centering
	\subtable[Comparison with different label generation methods.
		$^{\tau}$~means using the inference strategy of the image-level method.]{
		\small
		\setlength{\tabcolsep}{4.8mm}
		\begin{tabular}{lcccc}
			\toprule
			\ourdata$_{50}$          & mIoU & Img-Acc & $F_{\beta}$ \\ \midrule
			Image-level              & 26.9 & 57.6    & 53.0        \\
			Pixel-level              & 12.7 & 37.4    & 32.9        \\ \midrule
			Pixel-attention          & 29.3 & 65.5    & 49.0        \\
			Pixel-attention$^{\tau}$ & 29.2 & 61.7    & 52.3        \\
			\bottomrule
		\end{tabular}
		\label{tab:label-generation-method}
	}
	\subtable[Clustering time (second) of different label generation methods.]{
		\small
		\setlength{\tabcolsep}{1.5mm}
		\begin{tabular}{lccc}	\toprule
			                & \ourdata$_{50}$ & \ourdata$_{300}$ & \ourdata      \\ \midrule
			Image-level     & \tPow{2.8}{0}   & \tPow{8.9}{1}    & \tPow{7.5}{2} \\
			Pixel-level     & \tPow{3.2}{2}   & \tPow{4.6}{4}    & \tPow{4.1}{5} \\ \midrule
			Pixel-attention & \tPow{2.8}{0}   & \tPow{8.9}{1}    & \tPow{7.5}{2} \\
			\bottomrule
		\end{tabular}
		\label{tab:clustertime}
	}
	\subtable[Shared/unshared pixel-attention maps for the output features.]{
		\small
		\setlength{\tabcolsep}{5mm}
		\begin{tabular}{lccc}
			\toprule
			\ourdata$_{50}$ & mIoU & Img-Acc & $F_{\beta}$ \\ \midrule
			Shared          & 28.4 & 64.3    & 48.8        \\
			Unshared        & 29.3 & 65.5    & 49.0        \\
			\bottomrule
		\end{tabular}
		\label{tab:channel_pixelatt}
	}
	\subtable[Effectiveness of the fine-tuning step in our LUSS method.]{
		\small
		\setlength{\tabcolsep}{4.5mm}
		\begin{tabular}{lccc}
			\toprule
			\ourdata$_{50}$    & mIoU & Img-Acc & $F_{\beta}$ \\ \midrule
			Before fine-tuning & 26.0 & 63.8    & 44.7        \\
			After fine-tuning  & 29.3 & 65.5    & 49.0        \\
			\bottomrule
		\end{tabular}
		\label{tab:finetune-effect}
	}
\end{table}

\subsubsection{Label generation and Fine-tuning}
We evaluate the effectiveness of the proposed pixel-attention-based label generation
and fine-tuning scheme using the fully unsupervised evaluation protocol as described in~\secref{sec:protocols}.
We conduct ablation on the \ourdata$_{50}$ set unless otherwise stated.

\myPara{Effect of pixel-label generation.}
We compare our proposed pixel-attention-based pixel-label generation method
with image-level and pixel-level label generation methods.
We briefly introduce the label generation and fine-tuning process using
image-level and pixel-level label generation methods, respectively.
The image-level label generation method clusters $C$ categories over the pooled image-level embeddings
and assign image-level labels to each image.
During fine-tuning, a fully connected (FC) layer is supervised with image-level labels.
The FC layer is replaced with a 1$\times$1 conv layer to
obtain pixel-level segmentation masks during inference.
Due to lacking the `other' category,
we apply the class activation mapping (CAM) based mask generation method that is
widely used by WSSS methods to generate the final segmentation masks.
The implementation details are introduced in the supplementary.
Clustering on pixel-level embeddings is too costly on the large-scale \ourdata~dataset.
Instead, we implement the pixel-level method on the \ourdata$_{50}$ set for comparison.
We cluster $C+1$ categories using pixel-level embeddings and
fine-tune them with the pixel-level labels.
As shown in~\tabref{tab:label-generation-method},
the proposed pixel-attention-based label generation method outperforms
image-level and pixel-level methods
with a considerable margin.
The image-level method has better $F_\beta$ than our method.
We apply this inference strategy in our method,
and the $F_\beta$ is also significantly
improved while the mIoU has negligible change.

\myPara{Clustering time comparison.}
We compare the clustering time of pixel-attention-based label generation
with the other two label generation methods in~\tabref{tab:clustertime}.
Our method has the same clustering time as the image-level method
as they both use image-level embeddings.
Using output feature maps with the low-resolution of 7$\times$7,
the pixel-level method is much slower than our method
due to the huge number of pixels in the
training set.
When clustering on the full \ourdata~dataset,
the time of the pixel-level method is about \num{ 114 } hours,
which is unacceptable for real usage.

\myPara{Shared/unshared pixel-attention for output features.}
By default, we generate a unique pixel-attention map for each channel of output features.
We also study the effect of using one shared pixel-attention map for all channels.
The results in~\tabref{tab:channel_pixelatt} indicate
that using an unshared pixel-attention map for each channel results in better performance.
We visualize the pixel-attention maps of different channels in~\figref{fig:att_vis}.
Most channels focus on the semantic regions, while a few channels highlight
the background regions.
Also, the focus of each pixel-attention map is not identical,
explaining the effectiveness of unshared pixel-attention.

\myPara{Effect of fine-tuning.}
Our pixel-attention-based label generation method directly generates pixel-level segmentation masks, \ie pixel-level labels.
We compare the performance before/after the fine-tuning step
to validate the effect of fine-tuning step in our LUSS method.
As shown in~\tabref{tab:finetune-effect},
fine-tuning improves the test mIoU by 3.3\%, indicating that the generated pixel-level labels are still noisy
and fine-tuning further improves the semantic segmentation quality.

\begin{table}[t]
	\centering
	\small
	\caption{
		Transfer learning comparison among unsupervised representation learning methods pre-trained on \ourdata$_{300}$ and \ourdata~datasets.
		All models are trained with \num{ 100 } epochs. 		PASS$_{s/p}$ denotes using
		SwAV~\cite{caron2020unsupervised} and PixelPro~\cite{xie2020propagate}
		as the $L_e$ in~\eqref{eq:sumloss}, respectively.
		Supervised means initializing the model with image-level supervised pre-training.}
	\setlength{\tabcolsep}{0.56mm}
	\begin{tabular}{llcccccccc}
		\toprule
		\tRows{Transfer learning} & \thCols{COCO SEG} & \thCols{COCO DET} & VOC SEG                             \\ \cline{2-8}
		                          & AP                & AP50              & AP75    & AP   & AP50 & AP75 & mIoU \\ \midrule
		\multicolumn{8}{c}{\ourdata$_{300}$}                                                                    \\ \midrule
		Supervised                & 34.7              & 55.3              & 37.0    & 38.4 & 58.1 & 42.0 & 72.6 \\ \midrule
		Contrastive                                                                                             \\  \midrule
		\SimCLR                   & 31.9              & 51.1              & 34.1    & 35.0 & 53.7 & 38.2 & 66.4 \\
		\MoCovT                   & 33.7              & 53.6              & 36.1    & 37.1 & 56.3 & 40.3 & 67.8 \\
		\AdCo                     & 34.3              & 54.3              & 36.7    & 37.9 & 57.2 & 41.5 & 70.0 \\ \midrule
		Non-contrastive                                                                                         \\ \midrule
		\BYOL                     & 32.1              & 51.6              & 34.2    & 35.1 & 54.2 & 38.2 & 65.8 \\
		\SimSiam                  & 33.7              & 53.3              & 36.2    & 36.9 & 56.0 & 40.3 & 61.1 \\ \midrule
		Clustering                                                                                              \\ \midrule
		\PCL                      & 34.3              & 54.4              & 36.9    & 37.8 & 57.0 & 41.3 & 69.6 \\
		\SwAV                     & 32.4              & 52.1              & 34.6    & 35.5 & 54.9 & 38.6 & 68.9 \\
		\OursS                    & 33.8              & 53.7              & 36.2    & 37.2 & 56.6 & 40.6 & 70.8 \\
		\midrule
		Pixel-level                                                                                             \\ \midrule
		\DenseCL                  & 33.7              & 53.4              & 36.2    & 37.0 & 56.2 & 40.4 & 67.7 \\
		\PixelPro                 & 34.7              & 54.8              & 37.2    & 38.2 & 57.5 & 41.7 & 72.8 \\
		\OursP                    & 35.7              & 56.6              & 38.3    & 39.4 & 59.1 & 43.1 & 75.1 \\ 		\midrule
		\multicolumn{8}{c}{\ourdata}                                                                            \\
		\midrule
		Supervised                & 36.6              & 57.5              & 39.4    & 40.3 & 60.5 & 44.0 & 76.4 \\
		\SwAV                     & 34.4              & 55.0              & 36.8    & 37.8 & 58.0 & 41.1 & 73.0 \\
		\OursS                    & 35.3              & 56.0              & 37.8    & 38.9 & 58.8 & 42.3 & 75.3 \\
		\PixelPro                 & 35.9              & 56.6              & 38.6    & 39.5 & 59.2 & 43.1 & 73.9 \\
		\OursP                    & 36.5              & 57.4              & 39.1    & 40.2 & 60.3 & 44.1 & 76.1 \\
		\bottomrule
	\end{tabular}
	\label{tab:benchmarks-downstream}
\end{table}

\begin{table*}[t]
	\centering
	\setlength{\tabcolsep}{2.9mm}
	\caption{Ablation of shipping WSSS methods to the LUSS task.
		Properties in WSSS, \ie supervised pre-trained models, image-level GT labels, and large networks,
		that are not applicable in LUSS, make WSSS methods have a large performance drop in the LUSS task.}
	\begin{tabular}{lcccccccccccc}
		\toprule
		\tRows{\ourdata$_{50}$} & \tRows{Arch.}                 & \tRows{Param./MACC}
		                        & \tRows{Pre-train}             & \tRows{Labels}      & \tCols{mIoU}          & \tCols{Img-Acc}
		                        & \tCols{$F_{\beta}$}                                                                                         \\
		\multicolumn{5}{c}{}    & val                           & test                & val                   & test            & val  & test \\ \midrule
		\multirow{6}*{SEAM\cite{Wang_2020_CVPR}}
		                        & ResNet-38~\cite{resnet38d}    & 105.5M/100.4G       & Sup. ImageNet$_{1k}$  & GT
		                        & 49.7                          & 49.6                & 96.6                  & 95.7            & 61.5 & 60.9 \\
		                        & ResNet-18~\cite{He_2016_CVPR} & 11.3M/1.9G          & Sup. ImageNet$_{1k}$  & GT
		                        & 45.2                          & 44.5                & 90.9                  & 90.4            & 55.9 & 54.5 \\
		                        & ResNet-18~\cite{He_2016_CVPR} & 11.3M/1.9G          & Sup. ImageNet-50      & GT
		                        & 35.1                          & 35.8                & 81.2                  & 81.5            & 46.3 & 46.5 \\
		                        & ResNet-18~\cite{He_2016_CVPR} & 11.3M/1.9G          & MoCo. \ourdata$_{50}$ & -
		                        & 19.0                          & 19.1                & 45.1                  & 46.7            & 45.1 & 45.3 \\
		                        & ResNet-18~\cite{He_2016_CVPR} & 11.3M/1.9G          & SwAV. \ourdata$_{50}$ & -
		                        & 22.1                          & 22.3                & 54.6                  & 53.5            & 41.1 & 41.1 \\	\midrule
		\multirow{4}*{SC-CAM\cite{Chang_2020_CVPR}}
		                        & ResNet-18\cite{He_2016_CVPR}  & 11.5M/1.8G          & Sup. ImageNet$_{1k}$  & GT
		                        & 38.5                          & 39.3                & 81.9                  & 83.8            & 49.4 & 49.6 \\
		                        & ResNet-18\cite{He_2016_CVPR}  & 11.5M/1.8G          & Sup. ImageNet$_{50}$  & GT
		                        & 31.3                          & 32.1                & 70.2                  & 71.0            & 44.1 & 44.4 \\
		                        & ResNet-18~\cite{He_2016_CVPR} & 11.5M/1.8G          & MoCo. \ourdata$_{50}$ & -
		                        & 17.7                          & 18.1                & 43.7                  & 45.7            & 39.7 & 40.0 \\
		                        & ResNet-18~\cite{He_2016_CVPR} & 11.5M/1.8G          & SwAV. \ourdata$_{50}$ & -
		                        & 19.0                          & 19.7                & 50.0                  & 49.1            & 38.6 & 40.8 \\ \midrule
		\multirow{4}*{\begin{tabular}[c]{@{}l@{}}SEAM\cite{Wang_2020_CVPR}\\ +AdvCAM~\cite{lee2021antiadversarially}\end{tabular}}
		                        & ResNet-18~\cite{He_2016_CVPR} & 11.3M/1.9G          & Sup. ImageNet$_{1k}$  & GT
		                        & 46.9                          & 46.2                & 90.9                  & 90.4            & 58.4 & 57.5 \\
		                        & ResNet-18~\cite{He_2016_CVPR} & 11.3M/1.9G          & Sup. ImageNet$_{50}$  & GT
		                        & 36.9                          & 37.6                & 81.2                  & 81.5            & 49.2 & 49.6 \\
		                        & ResNet-18~\cite{He_2016_CVPR} & 11.3M/1.9G          & MoCo. \ourdata$_{50}$ & -
		                        & 19.2                          & 19.5                & 45.1                  & 46.7            & 46.8 & 47.3 \\
		                        & ResNet-18~\cite{He_2016_CVPR} & 11.3M/1.9G          & SwAV. \ourdata$_{50}$ & -
		                        & 23.7                          & 23.3                & 54.6                  & 53.5            & 44.2 & 43.9 \\ \bottomrule
	\end{tabular}
	\label{tab:weak_un_sss}
\end{table*}

\subsection{Transfer Learning to Other Tasks}
\label{sec:transfer}
Before the proposed LUSS task,
unsupervised representation learning methods mostly served as pre-training schemes
for transfer learning on downstream tasks~\cite{He_2020_CVPR,xie2020propagate}.
The LUSS task requires shape-related and category-related representations from self-supervised representation learning.
In this section, we study if the representation learned for the LUSS task
benefits the pixel-level downstream tasks, \eg semantic segmentation, instance segmentation, and object detection.
We also compare the effects of representation learning methods on LUSS and downstream tasks.
For fair comparisons, the ResNet-50~\cite{He_2016_CVPR} network is pre-trained on the
\ourdata$_{300}$ or \ourdata~datasets with \num{ 100 } epochs using
different representation learning methods unless otherwise stated.

\myPara{Instance segmentation and object detection.}
We utilize MaskRCNN~\cite{he2017mask} as the detector for instance segmentation
and object detection tasks.
Models are trained on the COCO17~\cite{lin2014microsoft} training set
and evaluated on the validation set.
Following common settings~\cite{xie2020propagate,he2017mask,He_2020_CVPR},
we load the weights of ResNet-50 pre-trained on different representation learning methods
and apply the 1$\times$ training schedule.
As shown in~\tabref{tab:benchmarks-downstream},
we validate our proposed learning strategies, \ie non-contrastive P2P alignment and D2S supervision
based on the SwAV~\cite{caron2020unsupervised} and PixelPro~\cite{xie2020propagate}.
We first compare models pre-trained on the \ourdata$_{300}$ dataset.
On instance segmentation, our method improves the SwAV and PixelPro
by 1.4\% and 1.0\% in mAP, respectively.
Similarly, the performance gain of our method over
SwAV and PixelPro are 1.7\% and 1.2\% in mAP on object detection, respectively.
These results prove that our representation learning method for the LUSS task has constant performance gains over different baselines
on instance segmentation and object segmentation tasks.
The pixel-level method PixelPro outperforms other image-level methods, \eg SwAV, AdCo, and SimSiam,
proving that pixel-level methods have a stronger transferring ability to these two pixel-level downstream tasks.
When pre-trained on the full \ourdata~dataset, our method still outperforms baselines,
\eg PixelPro based method has gains of 0.6\% and 0.7\% in mAP on
instance segmentation and object detection tasks, respectively.

\myPara{Semantic segmentation.}
We also transfer pre-trained models to the semantic segmentation task on the PASCAL VOC dataset~\cite{Everingham2009ThePV},
using ResNet-50 based Deeplab V3+~\cite{Chen_2018_ECCV} network.
The network is trained on the Pascal VOC SBD training set~\cite{BharathICCV2011} and evaluated on the validation set.
Following the training setting of~\cite{mmseg2020}, we train the network for 20k iterations with a batch size of \num{16}.
The images are scaled with a ratio of 0.5 to 2.0 and then cropped to \num{ 512 } for training.
When pre-trained on the \ourdata$_{300}$ dataset,
our method outperforms SwAV and PixelPro baselines by 1.9\% and 2.3\% in mIoU, respectively.
The performance gains over baselines are 2.3\% and 2.2\% in mIoU using the \ourdata~pre-trained models.
Pixel-level method PixelPro has a clear advantage over other image-level methods,
showing the pixel-level representation is crucial for semantic segmentation.
The contrast-based methods are better than clustering and non-contrastive methods for semantic segmentation
though they are both image-level methods.

\myPara{Relation between LUSS and transfer learning.}
We compare representation learning methods on the LUSS and downstream tasks
in~\tabref{tab:benchmarks} and~\tabref{tab:benchmarks-downstream}, respectively.
Compared among image-level methods,
the clustering method SwAV
has better performance on the LUSS task due to the high category accuracy.
On downstream tasks, SwAV is inferior to many methods
that achieve worse performance on the LUSS task.
For example, on the downstream instance segmentation task,
the contrastive method MoCov2 has a gain of 1.3\% in mAP over SwAV
but has a 10\% gap in mIoU on the LUSS task.
This observation is constant with the finding~\etal~\cite{kotar2021contrasting}
that contrastive methods learn better low-level features to benefit pixel-level downstream tasks.
Compared to image-level methods,
the pixel-level method PixelPro has a clear advantage over image-level methods.
But its performance is worse than many image-level methods on the LUSS task.
The pixel-level methods learn distinguishable pixel-level representations for downstream tasks
but lack enough category-related representations for the LUSS task.
Comparing methods within one category,
most of the well-performed methods on the LUSS task achieve better performance on downstream tasks.
Therefore, the LUSS and downstream tasks require different representations
but all benefit from high-quality representations.
We also demonstrate the effectiveness of our proposed P2P alignment and D2S supervision on
the LUSS task (\tabref{tab:p2p-d2s-luss}) and downstream tasks (\tabref{tab:p2p-d2s-downstream}).
Both proposed strategies improve the performance on LUSS and downstream tasks,
showing the generalizability of the proposed representation learning method.

\subsection{LUSS vs. WSSS}
\label{section_wsss}
Weakly supervised semantic segmentation (WSSS) with
image-level labels learn to segment semantic objects with only image-level labels.
We analyze the influence of typical settings in WSSS methods on the \ourdata$_{50}$ dataset,
\eg supervised ImageNet$_{1k}$ pre-trained models~\cite{ahn2018learning,ahn2019weakly,fan2020cian,sun2020mining,Chang_2020_CVPR},
image-level GT labels~\cite{shimoda2019self,sun2020mining}, and large network architectures~\cite{shimoda2019self,Wang_2020_CVPR,Chang_2020_CVPR},
and we show that these typical settings hinder shifting WSSS methods to the LUSS task.
Unless otherwise stated, the settings of WSSS methods are kept the same as official settings.
In unsupervised settings where GT labels are not available,
the self-generated image-level pseudo labels are utilized
to replace the GT labels in WSSS methods.

\begin{table}[t]
	\centering
	\setlength{\tabcolsep}{2.56mm}
	\caption{Semi-supervised semantic segmentation
		(semi-supervised evaluation protocol)
		using the \ourdata$_{50}$/\ourdata~datasets.
		PASS$_{s/p}$ denotes using
		SwAV~\cite{caron2020unsupervised} and PixelPro~\cite{xie2020propagate}
		as the $L_e$ in~\eqref{eq:sumloss}, respectively.
		Supervised means initializing the model with image-level supervised pre-training.
	}
	\begin{tabular}{lcccccccc}
		\toprule
		\tRows{Semi-supervised} & \tCols{mIoU} & \tCols{Img-Acc} & \tCols{F$_{\beta}$}                      \\ \cline{2-7}
		                        & val          & test            & val                 & test & val  & test \\	\midrule
		\multicolumn{7}{c}{\ourdata$_{300}$}                                                                \\ \midrule
		Supervised              & 27.7         & 27.5            & 61.1                & 62.3 & 64.3 & 64.9 \\
		\SimCLR                 & 12.7         & 12.6            & 34.4                & 34.8 & 59.1 & 59.6 \\
		\BYOL                   & 10.5         & 10.6            & 30.1                & 30.5 & 58.5 & 59.0 \\
		\MoCovT                 & 12.6         & 12.3            & 33.0                & 32.5 & 59.2 & 59.4 \\
		\DenseCL                & 16.2         & 16.0            & 34.9                & 35.7 & 61.0 & 60.9 \\
		\AdCo                   & 19.6         & 19.6            & 45.4                & 45.4 & 63.8 & 63.8 \\
		\PCL                    & 17.3         & 17.4            & 41.7                & 41.8 & 61.7 & 61.9 \\
		\SwAV                   & 23.0         & 23.3            & 51.2                & 51.5 & 64.0 & 64.0 \\
		\OursS                  & 25.7         & 25.7            & 52.3                & 52.8 & 65.5 & 66.0 \\
		\PixelPro               & 23.3         & 23.4            & 49.0                & 48.9 & 66.0 & 66.6 \\
		\OursP                  & 29.7         & 29.8            & 56.9                & 56.9 & 68.1 & 68.5 \\
		\midrule
		\multicolumn{7}{c}{\ourdata}                                                                        \\
		\midrule
		Supervised              & 25.7         & 25.0            & 57.3                & 57.4 & 66.3 & 66.7 \\
		\PixelPro               & 16.0         & 15.6            & 36.0                & 36.2 & 66.2 & 66.5 \\
		\OursP                  & 18.9         & 18.6            & 40.9                & 41.3 & 68.0 & 68.4 \\
		\SwAV                   & 18.2         & 17.9            & 42.8                & 43.2 & 66.0 & 66.2 \\
		\OursS                  & 19.4         & 19.2            & 43.3                & 43.4 & 66.6 & 66.9 \\
		\bottomrule
	\end{tabular}
	\label{tab:semi_proto}
\end{table}

\myPara{Pre-trained models.}
One of the main challenges in LUSS is to learn effective representations without supervision.
However, the effect of representation learning, \ie using weights pre-trained with different approaches,
is less explored in the WSSS methods.
Existing WSSS methods mostly utilize supervised ImageNet$_{1k}$ pre-trained models and fine-tune models on the semantic segmentation dataset~\cite{ahn2018learning,ahn2019weakly,fan2020cian,sun2020mining,Chang_2020_CVPR}, \eg PASCAL VOC~\cite{Everingham15}.
To understand the importance of pre-training,
we use different pre-trained models for SEAM~\cite{Wang_2020_CVPR}, SC-CAM~\cite{Chang_2020_CVPR},
and AdvCAM~\cite{lee2021antiadversarially}, as shown in~\tabref{tab:weak_un_sss}.
We observe that replacing the supervised ImageNet$_{1k}$ with the supervised ImageNet$_{50}$ dataset in SEAM~\cite{Wang_2020_CVPR} reduces the test mIoU
from 44.5\% to 35.8\%.
Replacing the supervised models with unsupervised models, \ie MoCo and SwAV, further reduces the test mIoU to 19.1\% and 22.3\%, respectively.
Both SC-CAM and AdvCAM suffered from the same issue,
indicating WSSS methods rely heavily on supervised pre-training.
The lack of supervised pre-training makes
the representation learning crucial to the LUSS task.
And our \ourdata~dataset provides a basis for fairly evaluating the representation quality of pre-trained models.

\myPara{Image-level GT labels.}
One essential difference between WSSS and LUSS tasks is that WSSS requires image-level GT labels.
Class activation maps~\cite{zhou2016learning,selvaraju2017grad}, commonly treated as the initial segment regions,
usually cover the most discriminative small area of objects.
Numerous WSSS methods heavily rely on GT labels to extend the CAM region to the whole object and remove the wrong region~\cite{sun2020fixing}
by image erasing~\cite{singh2017hide,li2018tell,wei2017object,hou2018self}, regions growing~\cite{kolesnikov2016seed,huang2018weakly,wang2018weakly,21TPAMI-OAA,jiang2022l2g},
stochastic feature selection~\cite{zhang2018adversarial,Lee_2019_CVPR},
gradients manipulation~\cite{lee2021antiadversarially},
or dataset level information~\cite{21PAMI_InsImgDatasetWSIS}.
To analyze the effect of GT labels on WSSS methods, we apply the recent work AdvCAM~\cite{lee2021antiadversarially}
to SEAM~\cite{Wang_2020_CVPR}.
AdvCAM anti-adversarially refines the CAM results by perturbing
the images along pixel gradients according to GT labels.
\tabref{tab:weak_un_sss} shows AdvCAM using GT labels
improves the baseline of ImageNet$_{1k}$ and ImageNet$_{50}$ pre-trained models with 1.7\% and 1.8\% in test mIoU.
However, when using generated pseudo labels and MoCo pre-trained model, the performance gain is only 0.4\%.
Using the SwAV pre-trained model with better image-level accuracy,
AdvCAM improves the model performance by 1.0\%.
Similarly, the SEAM and SC-CAM with GT labels outperform the unsupervised settings with a large margin.
Thus, the GT label reliance makes WSSS methods unable
to be directly shipped to the LUSS task due to the absence of the image-level GT label.

\myPara{Network architectures.}
Numerous network architectures have been developed to improve WSSS, including multi-scale enhancement~\cite{wei2018revisiting}
and affinity prediction~\cite{ahn2018learning,ahn2019weakly,fan2020cian}.
Due to the small size of the PASCAL VOC dataset,
many state-of-the-art WSSS methods improve the performance using large models
with extensive parameters and computational cost,
\eg wide ResNet-38~\cite{resnet38d,shimoda2019self,Wang_2020_CVPR,Chang_2020_CVPR}
and ResNet with small output strides~\cite{chen2020weakly,lee2021antiadversarially}.
As the proposed \ourdata~datasets are \num{ 44 } to \num{ 800 } times larger than PASCAL VOC, the computational cost of training LUSS models with large models used by WSSS methods is prohibitively high.
To analyze the effect of model architectures,
we change the network in SEAM~\cite{Wang_2020_CVPR} (see ~\tabref{tab:weak_un_sss}).
We remove the Deeplab re-training step used in WSSS methods for fair comparisons.
When replacing the ResNet-38~\cite{resnet38d} with a standard ResNet-18~\cite{He_2016_CVPR},
the test mIoU drops from 49.6\% to 44.5\%.
Large models benefit the performance,
but the high computational cost makes the
unsupervised training of LUSS models impracticable.

\begin{table}[t]
	\centering
	\small
	\setlength{\tabcolsep}{3mm}
	\caption{mIoU results of supervised backbone models using distance matching evaluation protocol on
		the \ourdata~testing set. Top-1 Acc. is the classification accuracy on the \ourdata~testing set.$^{*}$ indicates models are finetuned with the ImageNet-S semi-supervised segmentation training set.}
	\begin{tabular}{lcccccc}
		\toprule
		Supervised                                            & Top-1 Acc. & mIoU \\
		\midrule
		\thCols{\ourdata}                                                         \\
		\midrule
		ResNet-50\cite{He_2016_CVPR}                          & 83.6       & 29.8 \\
		ResNet-101\cite{He_2016_CVPR}                         & 84.3       & 31.4 \\
		DenseNet-161\cite{huang2017densely}                   & 84.3       & 29.8 \\
		Inception V3\cite{Szegedy_2016_CVPR}                  & 77.7       & 29.9 \\
		ResNeXt-50\cite{xie2017aggregated}                    & 84.4       & 32.6 \\
		ResNeXt-101\cite{xie2017aggregated}                   & 85.5       & 34.8 \\
		EfficientNet-B3\cite{efficientnet}                    & 85.3       & 32.3 \\
		Res2Net-50\cite{gao2019res2net}                       & 84.8       & 35.7 \\
		Res2Net-101\cite{gao2019res2net}                      & 85.6       & 37.2 \\
		Swin-S\cite{Liu_2021_ICCV}                            & 87.8       & 38.6 \\
		Swin-B\cite{Liu_2021_ICCV}                            & 88.0       & 38.2 \\ \midrule
		ConvNeXt-T$^{*}$~\cite{liu2022convnet}                & -          & 45.1 \\
		RF-ConvNeXt-T$^{*}$ (SingleRF)~\cite{gao2022rfnext}   & -          & 46.2 \\
		RF-ConvNeXt-T$^{*}$ (MultipleRF)~\cite{gao2022rfnext} & -          & 47.0 \\
		\bottomrule
	\end{tabular}
	\label{tab:cam_evaluation}
\end{table}

\subsection{Applications of the \ourdata~dataset}
The proposed \ourdata~dataset has pixel-level annotations,
thus has more applications apart from the LUSS task.
This section presents the \ourdata~dataset
for the large-scale semi-supervised semantic segmentation,
evaluation of image-level supervised backbone models,
and salient object detection with a subset.

\myPara{Large-scale semi-supervised semantic segmentation.}
Semi-supervised semantic segmentation requires training with a small part of labeled data
and many unlabeled data.
Fine-tuning trained LUSS models on the 1\% labeled training images of the \ourdata~dataset
achieves the semi-supervised semantic segmentation,
which is the semi-supervised evaluation protocol of LUSS as introduced in~\secref{sec:protocols}.
We follow the training scheme of fine-tuning step in~\secref{sec:training_detail},
except that models are trained with \num{ 30 } epochs using GT labels.
The semi-supervised semantic segmentation results are shown in~\tabref{tab:semi_proto}.
Our proposed method outperforms the SwAV and PixelPro baselines with considerable margins on
\ourdata$_{300}$ and \ourdata~datasets,
respectively.
Our PixelPro based method even suppresses the image-level supervised model on the \ourdata$_{300}$
dataset.
In the semi-supervised setting, PixelPro has a similar performance to SwAV,
but SwAV has a large advantage over PixelPro
in distance matching evaluation results (see~\tabref{tab:benchmarks}).
We conclude that fine-tuning models with pixel-level GT labels
make models require less self-learned category-related representation ability.

\myPara{Evaluate supervised backbone models.}
Apart from the LUSS task,
the \ourdata~dataset can also evaluate the shape and category representation ability
of backbone models trained with image-level supervision.
We benchmark the mIoU of backbone models on the \ourdata~testing set using distance matching evaluation protocol,
as shown in~\tabref{tab:cam_evaluation}.
As a reference, we also obtain the top-1 classification accuracy of these models on the \ourdata~dataset.
We observe that image-level top-1 accuracy is not always constant with the mIoU,
indicating that models with good category representation might not be good at shape representation.
To observe how much the ImageNet-S
dataset can benefit from a good backbone model,
we test the recent proposed RF-ConvNeXt~\cite{gao2022rfnext}
that enhances the ConvNeXt~\cite{liu2022convnet}
with more suitable receptive fields.
RF-ConvNeXt achieves high semantic segmentation performance,
indicating a good backbone network
is needed for the ImageNet-S semantic segmentation.

\myPara{Salient object detection with ImageNet-S subset.}
Salient object detection (SOD) aims at segmenting salient objects
regardless of categories~\cite{ChengPAMI}.
Due to the category insensitive property of
SOD~\cite{21PAMI-Sal100K},
an unsupervised SOD model can provide
shape prior knowledge to LUSS models.
To facilitate the SOD task under large-scale data,
we construct a SOD dataset, namely ImageNet-Sal, by
selecting images with salient objects from the ImageNet-S dataset.
For pixel-level labeled images in train/val/test sets of ImageNet-S,
we manually select images with salient objects and remove the no-salient annotations.
For unlabeled images in the training set,
we pick salient images with the help of several pre-trained SOD models.
As the picked images might not contain salient objects,
we encourage future SOD methods to self-identify training images
with salient objects.

\section{Conclusions}
\label{sec:conclusion}
This work proposes a new problem of large-scale unsupervised semantic segmentation
to facilitate semantic segmentation in real-world environments with a large diversity and large-scale data.
We present a benchmark for LUSS to provide large-scale data with high diversity,
a clear task objective, and sufficient evaluation.
We present one new method of LUSS to assign labels to pixels
with category and shape representations learned
from the large-scale data without human-annotation supervision.
The LUSS method contains enhanced representation learning and pixel-attention assisted pixel-level label generation strategy.
We evaluate our method with multiple evaluation protocols and
reveal the potential of LUSS to pixel-level downstream tasks, \eg semantic segmentation.
In addition, we benchmark unsupervised representation learning
methods and weakly supervised semantic segmentation methods,
and we summarize the challenges and possible directions of LUSS.

\myPara{Acknowledgement}
This work is funded by the National Key Research and Development Program
of China Grant No.2018AAA0100400,
NSFC (62225604),
and the Fundamental Research Funds for the Central Universities
(Nankai University, NO. 63223050).
Thanks for part of the pixel-level annotation from the Learning from Imperfect Data Challenge~\cite{lid2019}.

\section{Supplementary}
\subsection{Evaluation metrics}
We use the mean intersection over union (mIoU), image-level accuracy (Img-Acc),
and F-measure ($F_{\beta}$) as the evaluation metrics for the LUSS task.
During the evaluation, all images are evaluated with the original image resolution.
We give the implementation details of these evaluation metrics.

\myPara{Mean IOU.}
Similar to the supervised semantic segmentation task
\cite{Everingham15,zhou2018semantic},
we utilize the mIoU metric
to evaluate the segmentation mask quality.
Suppose the images of the validation/testing sets have $N^{P}$ pixels.
Pixels have GT labels $\mathbf{G}^{P}=\{\mathbf{G}^{P}_k, k \in [1, N^{P}] \}$
and predicted labels $\mathbf{P}^{P}=\{\mathbf{G}^{P}_k, k \in [1, N^{P}] \}$,
where $\mathbf{G}^{P}_k$ and $\mathbf{P}^{P}_k$ are the GT and
predicted label of the $k$-th pixel in the validation/testing sets.
We use $t_{ij}$ to represent the number of pixels that belong to
GT category $i$ and are predicted as category $j$,
which is written as:
\begin{equation}
	t_{ij}=\sum_{k=1}^{N^{P}}\mathbb I \{\mathbf{G}^{P}_k=i~\&~\mathbf{P}^{P}_k=j\}.
\end{equation}
Then, the mIoU is calculated as:
\begin{equation}
	{\rm mIoU}=\frac{1}{C + 1}\sum_{i=0}^{C}\frac{t_{ii}}{\sum_{j=0}^{C}t_{ij} + \sum_{j=0}^{C}t_{ji}-t_{ii}},
\end{equation}
where $0$-th category is the `other' category,
which is also considered to get mIoU.

\myPara{Image-level accuracy.}
The Img-Acc can evaluate the category representation ability of models.
As many images contain multiple labels,
we follow~\cite{beyer2020we} and treat the predicted label as correct
if the predicted category with the largest area is within the GT label list.
Suppose the validation/testing sets have $N$ images and $C$ categories.
We present the image set as $\mathbf{D}=\{\mathbf{D}_k, k \in [1,N] \}$ with ground-truth (GT) labels $\mathbf{G}=\{ \mathbf{G}_k, k \in [1,N]\}$
and predicted labels $\mathbf{P}=\{\mathbf{P}_k, k \in [1,N] \}$,
where $\mathbf{G}_k$ and $\mathbf{P}_k$ are the GT and predicted category sets of the image $\mathbf{D}_k$.
The image-level accuracy is obtained as follows:
\begin{equation}
	Acc = \frac{1}{N} \sum_{k=1}^{N} \mathbb I \{ ~{\rm largest}(\mathbf{P}_k) \in \mathbf{G}_k \},
\end{equation}
where ${\rm largest}(\mathbf{P}_k)$ indicates the category with the largest area of the image,
and $\mathbb I $ is one when this category belongs to $\mathbf{G}_k$.

\myPara{F-measure.}
In addition to category-related representation,
we utilize $F_{\beta}$ to evaluate the shape quality,
which ignores the semantic categories.
Specifically, we treat pixels of major categories as the foreground category and pixels in the `other' category as the background category.
Suppose the image $i$ has $N_{i}$ pixels,
and the corresponding foreground GT mask is denoted as $\mathbf{G}^{F}_{i}=\{\mathbf{G}^{F}_{ik}, k \in [1, N_{i}] \}$
where the $\mathbf{G}^{F}_{ik}$ equals 1 when the label of the pixel does not belong to the `other' category.
The predicted foreground mask of image $i$, denoted by $\mathbf{P}^{F}_{i}=\{\mathbf{P}^{F}_{ik}, k \in [1, N_{i}] \}$, is obtained in the same way.
The ${\rm precision}_{i}$ of image $i$ is calculated as follows:
\begin{equation}
	{\rm precision}_{i}=\frac{\sum_{k=1}^{N_{i}}\mathbb I \{\mathbf{G}^{F}_{ik} \cdot \mathbf{P}^{F}_{ik} = 1\}}{\sum_{k=1}^{N_{i}}\mathbb I \{\mathbf{P}^{F}_{ik} = 1\}}.
\end{equation}
Then the ${\rm recall}_{i}$ of image $i$ is calculated as follows:
\begin{equation}
	{\rm recall}_{i}=\frac{\sum_{k=1}^{N_{i}}\mathbb I \{\mathbf{G}^{F}_{ik} \cdot \mathbf{P}^{F}_{ik} = 1\}}{\sum_{k=1}^{N_{i}}\mathbb I \{\mathbf{G}^{F}_{ik} = 1\}}.
\end{equation}
With ${\rm recall}$ and ${\rm precision}$ for each image, we get $F_{\beta}$ of the dataset as follows:
\begin{equation}
	F_{\beta} = \frac{1}{N}\sum_{i=1}^{N}\frac{(1+\beta^{2})  \cdot {\rm precision}_{i} \cdot {\rm recall}_{i}}{\beta^{2}  \cdot {\rm precision}_{i} + {\rm recall}_{i}},
\end{equation}
where $N$ is the number of images in the validation/testing set,
and we follow common settings to set $\beta^{2} = 0.3$~\cite{ChengPAMI}.

\renewcommand{\tRows}[1]{\multirow{2}*{#1}}
\renewcommand{\tCols}[1]{\multicolumn{2}{c}{#1}}
\def\MDC{MDC\cite{caron2018deep,cho2021picie}}
\def\PiCIE{PiCIE\cite{cho2021picie}}
\def\MaskCon{MaskCon\cite{van2020unsuperv}}

\begin{table*}[th]
	\caption{Comparison of our proposed LUSS method and existing
		USS methods on the \ourdata~dataset
		using fully unsupervised evaluation protocol.
		The `other' category is not considered during the evaluation.
	}
	\vspace{10pt}
	\centering
	\footnotesize
	\setlength{\tabcolsep}{2.0mm}
	\begin{tabular}{lccccccccccccccccc}
		\toprule
		\tRows{LUSS}    & \tRows{Pior}        & \tCols{mIoU}             & \tCols{b-mIoU}             &
		\tCols{Img-Acc} & \tCols{F$_{\beta}$} & \multicolumn{4}{c}{mIoU} & \multicolumn{4}{c}{b-mIoU}                                                                                                                                                                                       \\
		                &                     & val                      & test                       & val        & test       & val        & test       & val        & test       & S.         & M.S.       & M.L.       & L.         & S.         & M.S.       & M.L.       & L.         \\ \midrule
		\multicolumn{17}{c}{\ourdata$_{50}$}                                                                                                                                                                                                                                                \\ \midrule
		\MDC            & -
		                & 4.0                 & 3.7                      & 1.4                        & 1.2        & 14.9       & 13.4       & 31.6       & 31.3       & 0.4        & 2.6        & 3.8        & 5.0        & 0.2        & 1.1        & 1.4        & 1.5                     \\
		\MDC            & \textit{I}
		                & 14.9                & 14.6                     & 3.2                        & 3.1        & 44.8       & 40.8       & 33.2       & 32.6       & 2.7        & 11.1       & 14.93      & 19.51      & 0.9        & 2.2        & 3.2        & 4.8                     \\
		\PiCIE          & -
		                & 5.0                 & 4.6                      & 1.8                        & 1.6        & 15.8       & 14.0       & 14.6       & 32.2       & 0.2        & 3.1        & 5.1        & 5.4        & 0.1        & 1.2        & 1.7        & 1.9                     \\
		\PiCIE          & \textit{I}
		                & 18.1                & 17.9                     & 3.7                        & 4.0        & 45.0       & 44.0       & 32.1       & 31.6       & 4.4        & 13.3       & 20.4       & 23.5       & 0.9        & 2.7        & 4.4        & 5.8                     \\
		\MaskCon        & \textit{S}
		                & 23.5                & 23.1                     & 14.8                       & 14.3       & 47.9       & 47.6       & 65.7       & 66.2       & 10.2       & 24.4       & 23.7       & 19.7       & 8.9        & 16.1       & 13.7       & 9.9                     \\
		\MaskCon\dag    & \textit{S}
		                & 12.7                & 9.2                      & 7.6                        & 5.3        & 30.2       & 22.4       & 62.6       & 62.3       & 1.5        & 8.1        & 9.9        & 9.7        & 1.2        & 5.3        & 5.6        & 4.8                     \\
		\OursS          & -
		                & 28.4                & 28.6                     & 6.9                        & 6.7        & {\bf 66.2} & {\bf 65.5} & 49.0       & 49.0       & 4.6        & 24.1       & 32.5       & 32.4       & 1.9        & 5.4        & 7.4        & 8.9                     \\
		\OursP          & -
		                & 31.9                & 31.5                     & 6.6                        & 6.6        & 62.9       & 64.1       & 48.7       & 47.9       & 8.5        & 25.6       & 36.0       & 40.3       & 4.1        & 5.1        & 7.2        & 9.8                     \\
		\OursP + RC     & \textit{S}          & 42.0                     & 41.5                       & 16.8       & 17.1       & 58.8       & 61.8       & 62.1       & 61.3       & 15.8       & 37.9       & {\bf 44.8} & {\bf 43.2} & 10.3       & 16.5       & 18.3       & {\bf 16.4} \\
		\OursP  + Sal   & \textit{S}          & {\bf 42.5}               & {\bf 41.5}                 & {\bf 19.7} & {\bf 19.5} & 64.6       & 65.2       & {\bf 70.0} & {\bf 69.9} & {\bf 17.2} & {\bf 40.8} & 44.4       & 37.8       & {\bf 13.6} & {\bf 21.8} & {\bf 19.9} & 14.7       \\
		\midrule
		\multicolumn{17}{c}{\ourdata$_{300}$}                                                                                                                                                                                                                                               \\
		\midrule
		\OursP          & -
		                & 16.6                & 16.0                     & 4.4                        & 4.2        & 34.7       & 32.8       & 34.4       & 34.3       & 2.8        & 12.0       & 16.5       & 21.8       & 1.4        & 3.2        & 3.9        & 6.4                     \\
		\OursS          & -
		                & {\bf 17.9}          & {\bf 18.0}               & {\bf 5.1}                  & {\bf 5.1}  & {\bf 43.9} & {\bf 42.6} & {\bf 47.6} & {\bf 47.5} & {\bf 4.0}  & {\bf 13.4} & {\bf 19.4} & {\bf 23.5} & {\bf 1.9}  & {\bf 4.1}  & {\bf 5.4}  & {\bf 7.0}               \\
		\midrule
		\multicolumn{17}{c}{\ourdata}                                                                                                                                                                                                                                                       \\
		\midrule
		\OursP          & -
		                & 7.3                 & 6.6                      & 2.4                        & 2.1        & 19.9       & 18.0       & 34.8       & 34.6       & 1.3        & 4.6        & 7.1        & 8.4        & 0.6        & 1.5        & 2.1        & 2.8                     \\
		\OursS          & -
		                & {\bf 11.5}          & {\bf 11.0}               & {\bf 3.8}                  & {\bf 3.5}  & {\bf 24.0} & {\bf 22.3} & {\bf 37.1} & {\bf 36.9} & {\bf 2.4}  & {\bf 8.3}  & {\bf 11.9} & {\bf 13.4} & {\bf 1.2}  & {\bf 3.0}  & {\bf 3.7}  & {\bf 4.3}               \\
		\bottomrule
	\end{tabular}
	\label{tab:unsup_com_noother}
\end{table*}

\subsection{CAM-based Inference}
\label{sec:cam_infer}
We introduce the CAM-based segmentation mask inference strategy used in this work.
For each image, we have $A^{c} \in R^{H \times W}$ as the CAM for the predicted category $c$.
Then, the CAM is normalized as follows:
\begin{equation}
	\hat{A}^{c} = \frac{A^{c} - \min A^{c}}{\max A^{c} - \min A^{c}}.
\end{equation}
Then, we assign label $c$ to regions based on the activation values.
Specifically, the assigned label at the position $(x,y)$ is
\begin{equation}
	\mathbf{f_{(x,y)}} =
	\begin{cases}
		0 & \hat{A}^{k}(x,y) < \tau;   \\
		k & \hat{A}^{k}(x,y) \ge \tau.
	\end{cases}
\end{equation}
where $\tau$ indicates a threshold.

\subsection{Performance Without `other' Category}
Normally, predicting the `other' category is easier than
the rest since the `other' category covers
a larger area of images.
Therefore, it's possible that including the `other' category makes it easier for the model to get higher mIoU.
We evaluate the performance without the `other' category
in ~\tabref{tab:unsup_com_noother}.
We observe that the mIoU differences with/without the `other' category are mostly less
than $0.1\%$ on mIoU for the ImageNet-S$_{300/919}$ sets.
On the ImageNet-S$_{50}$ set,
there is about a $1\%$ gap in mIoU.
Since the mIoU is obtained by averaging IoU on all categories,
the `other' category has a small influence on
the mIoU.

\bibliographystyle{IEEEtran}
\bibliography{egbib}
\newcommand{\AddPhoto}[1]{{\includegraphics[width=1in,keepaspectratio]{figures/Authors/#1}}}
\newcommand{\AuthorBio}[3]{\vspace{-.2in}\begin{IEEEbiography}[\AddPhoto{#1}]{#2}#3\end{IEEEbiography}}

\AuthorBio{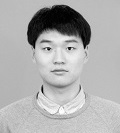}{Shanghua Gao}{
	is a Ph.D. candidate in Media Computing Lab at Nankai University.
	He is supervised via Prof. Ming-Ming Cheng.
	His research interests include computer vision and representation learning.
}

\AuthorBio{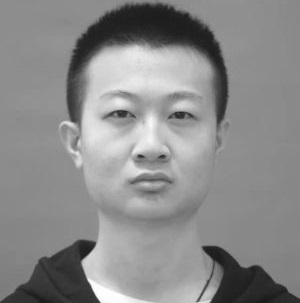}{Zhong-Yu Li}{
	is a Ph.D. student from the college of computer science, Nankai university.
	He is supervised via Prof. Ming-Ming cheng.
	His research interests include deep learning, machine learning and computer vision.
}

\AuthorBio{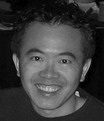}{Ming-Hsuan Yang}{
	is a professor in Electrical
	Engineering and Computer Science at University of California, Merced.
	He received the PhD degree in Computer Science from the University
	of Illinois at Urbana-Champaign in 2000. Yang
	has served as an associate editor of the IEEE TPAMI,
	IJCV, CVIU, \etc.
	He received the
	NSF CAREER award in 2012 and the Google Faculty Award in 2009.
}

\AuthorBio{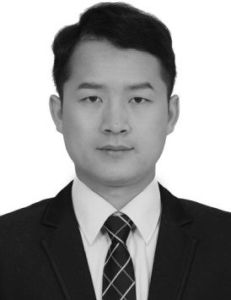}{Ming-Ming Cheng}{
	received his PhD degree from Tsinghua University in 2012,
	and then worked with Prof. Philip Torr in Oxford for 2 years.
	He is now a professor at Nankai University, leading the
	Media Computing Lab.
	His research interests includes computer vision and computer graphics.
	He received awards including ACM China Rising Star Award,
	IBM Global SUR Award, \etc.
	He is a senior member of the IEEE and on the editorial boards of
	IEEE TPAMI and IEEE TIP.
}

\AuthorBio{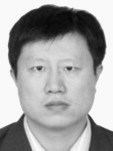}{Junwei Han}{
	is currently a Full Professor with
	Northwestern Polytechnical University, Xi’an,
	China. His research interests include computer
	vision, multimedia processing, and brain imaging analysis.
	He is an Associate Editor of IEEE Trans. on Human-Machine Systems,
	Neurocomputing, Multidimensional Systems and Signal
	Processing, and Machine Vision and Applications.
}

\AuthorBio{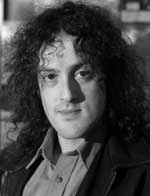}{Philip Torr}{
	received the PhD degree from Oxford University.
	After working for another three years at Oxford,
	he worked for six years for Microsoft Research, first in Redmond,
	then in Cambridge, founding the vision side of the Machine
	Learning and Perception Group.
	He is now a professor at Oxford University.
	He has won awards from top vision conferences,
	including ICCV, CVPR, ECCV, NIPS and BMVC.
	He is a senior member of the IEEE and Fellow of the Royal Society.
}

\vfill

\end{document}